\documentclass[preprint, 12pt]{elsarticle}

\usepackage{amsmath}
\usepackage{hyperref}
\usepackage{lineno}
\usepackage{float}
\usepackage{caption}
\usepackage{graphicx}
\usepackage{multicol}
\usepackage{pgfplots}
\usepackage{subcaption}
\usepackage{array}
\usepackage{booktabs}
\usepackage{multirow}
\usepackage{tikz}
\usetikzlibrary{bayesnet}
\usetikzlibrary{arrows}
\usepackage{color}
\usepackage{bm}
\usetikzlibrary{backgrounds}

\usepackage{array,graphicx}
\usepackage{booktabs}
\usepackage{pifont}
\usepackage{hyperref}
\usepackage{xr}
\newcommand*\rottwo{\rotatebox{90}}
\newcommand*\OK{\ding{51}}
\newcommand{\supp}[2]{#1^{(#2)}}

\journal{Forensic Science International}

\begin{document}

\begin{frontmatter}
  \title{Modeling human decomposition: a Bayesian approach}

  \author[inst1]{D.~Hudson Smith}
\author[inst2]{Noah Nisbet}
\author[inst3]{Carl Ehrett}
\author[inst4]{Cristina I.~Tica}
\author[inst5]{Madeline M.~Atwell}
\author[inst5]{Katherine E.~Weisensee}

\affiliation[inst1]{organization={Department of Mathematical and Statistical Sciences, Clemson University},
    addressline={220 Parkway Dr.},
    city={Clemson},
    postcode={29630},
    state={SC},
    country={USA},
}
\affiliation[inst2]{organization={School of Computing, Clemson University},
    city={Clemson},
    postcode={29630},
    state={SC},
    country={USA},
}
\affiliation[inst3]{organization={Research Computing and Data, Clemson University},
    city={Clemson},
    postcode={29630},
    state={SC},
    country={USA},
}
\affiliation[inst4]{organization={Department of Anthropology and Applied Archaeology, Eastern New Mexico University},
    city={Portales},
    postcode={88130},
    state={NM},
    country={USA},
}
\affiliation[inst5]{organization={Department of Sociology, Anthropology	and Criminal Justice, Clemson University},
    city={Clemson},
    postcode={29630},
    state={SC},
    country={USA},
}

  \begin{abstract}
    Environmental and individualistic variables affect the rate of human decomposition in complex ways. These effects complicate the estimation of the postmortem interval (PMI) based on observed decomposition characteristics. In this work, we develop a generative probabilistic model for decomposing human remains based on PMI and a wide range of environmental and individualistic variables. This model explicitly represents the effect of each variable, including PMI, on the appearance of each decomposition characteristic, allowing for direct interpretation of model effects and enabling the use of the model for PMI inference and optimal experimental design. In addition, the probabilistic nature of the model allows for the integration of expert knowledge in the form of prior distributions. We fit this model to a diverse set of 2,529 cases from the GeoFOR dataset. We demonstrate that the model accurately predicts 24 decomposition characteristics with an ROC AUC score of 0.85. Using Bayesian inference techniques, we invert the decomposition model to predict PMI as a function of the observed decomposition characteristics and environmental and individualistic variables, producing an R-squared measure of 71\%. Finally, we demonstrate how to use the fitted model to design future experiments that maximize the expected amount of new information about the mechanisms of decomposition using the Expected Information Gain formalism.
  \end{abstract}

  \begin{graphicalabstract}
    \begin{figure}[H]
      \centering
      \includegraphics[width=1.0\textwidth]{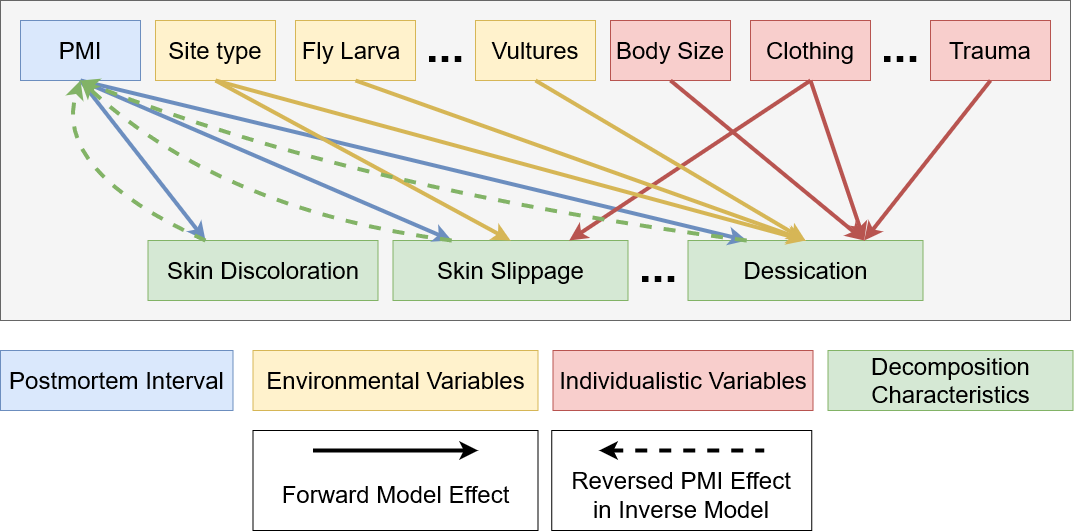}
      \caption{This graphic represents our generative model for decomposition based on a range of environmental and individualistic variables. Solid-line arrows represent the effect of one variable on the occurrence of a particular decomposition characteristic. A small subset of effects are allowed based on the stage of decomposition associated with each characteristic. The dashed-line arrows represent the use of the model to infer postmortem interval via Bayesian inference.}
      \label{fig:graphical_abstract}
    \end{figure}
  \end{graphicalabstract}

  \begin{highlights}
    \item We model 24 decomposition characteristics based on 18 variables
    \item We fit our model to 2,529 cases from the geoFOR dataset
    \item Our model predicts decomposition with ROC AUC of 0.85 and PMI with an $R^2$ of 71\%
    \item We present significant findings on a wide range of decomposition mechanisms    
    \item We demonstrate how to use the model for optimal experimental design
  \end{highlights}


\end{frontmatter}
\section{Introduction}

Environmental variables and individual characteristics influence the decomposition of human remains in nuanced, interacting ways. In addition, decomposition is a multifaceted concept with various traits that may or may not manifest at different stages of the decomposition process. A more thorough understanding of the effects of environmental and individual variables on the facets of decomposition would allow forensic investigators to make more detailed inferences about the circumstances surrounding unattended deaths. Despite this importance, most prior research relies on simplistic measures of both environmental factors and decomposition, such as accumulated degree days (ADD) and total body score (TBS) \cite{megyesi2005using}, or variations of the TBS formula in different environmental contexts, despite the fact that many report unsatisfactory results \cite{suckling2016longitudinal,ceciliason2018quantifying,giles2020effect,gunawardena2023retrospective,gelderman2018development,pittner2020applicability,wescott2018validation}. In addition, most studies use few cases collected in narrow environments, limiting the generalizability of the findings. As a result, the forensic science community lacks any principled, general-purpose models describing the effects of environmental and individual variables on the decomposition process. In the words of Suckling et al., ``The attempt to create a single equation that can provide a quantitative estimate of postmortem interval (PMI) from decomposition scores is laudable, but as this study shows, compromising environmental specificity sacrifices precision" \cite{suckling2016longitudinal}.

We begin to address this gap by developing a series of general-purpose models for the decomposition process. We leverage the standardized taphonomic characteristics and large case dataset from the geoFOR platform \cite{weisensee2024geofor} to develop these models. As shown in Fig.~\ref{fig:graphical_abstract}, our models explicitly quantify how 18 environmental and individual features affect the onset of 24 encompassing macromorphoscopically visual decomposition characteristics, as outlined in the geoFOR application. Given geoFOR's comprehensive set of postmortem physiological changes and taphonomic characteristics, these findings shed light on a diverse group of core forensic science questions, such as the impact of body size on the appearance of desiccation. Moreover, our statistical models accurately predict the manifestation of decomposition and the PMI for withheld cases.

We estimate our model parameters using the Bayesian probabilistic modeling formalism \cite{bishop2013model,bishop2006pattern}. The affordances of this approach make it ideal for expanding the science of human decomposition. Firstly, it allows us to directly incorporate prior scientific knowledge by excluding unrealistic effects from the model structure and specifying prior beliefs about effect sizes. Secondly, the Bayesian approach provides a principled procedure for making inferences about quantities of interest, such as PMI. Thirdly, this approach allows us to quantify the uncertainty in effect sizes and model predictions, such as the prediction of PMI for new cases. Lastly, the Bayesian formalism provides a principled approach to designing future experiments to maximize learning about specific decomposition mechanisms.

\section{Related work}
\paragraph{Physico-chemical models} Physico-chemical models of human decomposition provide a first-principles description of decomposition in terms of the aerobic and anaerobic processes that occur following death and the interactions of these processes with the human microbiome as well as foreign bacteria from soil \cite{dent2004review,statheropoulos2007environmental,vass1992time}. These models play a critical role in our understanding of human decomposition and motivate the development of data-driven methods for PMI prediction \cite{johnson2016machine, belk2018microbiome}. Though promising, these methods only apply within the early stages of decomposition. In addition, they face significant adoption challenges resulting from the need for validated, fieldable technologies for measuring microbiome levels and making predictions based on those measurements \cite{metcalf2019estimating}.

\paragraph{The Megyesi paradigm} In 2005, Megyesi et al. \cite{megyesi2005using} introduced the Total Body Score (TBS) metric based on the stages of decomposition described by Galloway et al. \cite{galloway1989decay}. In addition, Megyesi et al. introduced a statistical formula relating Accumulated Degree Days (ADD) to TBS based on 68 human cases. Unlike the physico-chemical approach, the Megyesi approach is easily adopted as it relies upon recording visual morphological changes on the postmortem body, rather than requiring expensive or time consuming technologies. The idea of modeling decomposition using TBS-like measurements became a paradigm for decomposition research and forensic practice. Despite significant conceptual flaws in the original work \cite{moffatt2016improved,smith2023commentary} and multiple demonstrations of the inaccuracy of the Megyesi formula \cite{wescott2018validation,suckling2016longitudinal,ceciliason2018quantifying,giles2020effect,wescott2018validation}, this method still holds significant sway over scientific study design and forensic practice. However, recent developments suggest a trend toward developing a more principled forensic taphonomy \cite{iqbal2020recent,miles2020review}.

\paragraph{geoFOR} The geoFOR platform\footnote{https://www.geoforapp.info/} allows forensic investigators to estimate PMI based on 24 decomposition characteristics along with variables describing the environment and subject \cite{weisensee2024geofor}. Users of the platform input structured, anonymized case information, and the platform provides a PMI prediction along with an uncertainty estimate using a machine learning model. Users can also upload the date that the individual was last known to be alive or more precise date of death information if known. The uploaded case data is then included in the geoFOR reference case dataset and will be available to researchers upon request. The geoFOR team uses this data to train the PMI prediction model. We leverage geoFOR's standardized knowledge structure and large case dataset to build our empirical taphonomic model. In the geoFOR manuscript \cite{weisensee2024geofor}, the authors present a machine-learning-based predictive model for PMI. This model directly predicts PMI based on the decomposition, environmental, and individual variables related to the body available in the geoFOR ontology. This approach gives the best PMI predictive power but does not directly shed light on decomposition mechanisms. In contrast, our Bayesian model describes the manifestation of the decomposition variables based on environmental and individual factors. Having thus modeled the decomposition process, we can then identify the PMI values that are consistent with the observed state of decomposition using Bayesian inference techniques.

\paragraph{Bayesian modeling} Bayesian modeling techniques have a somewhat controversial history in the forensic science community due to questions surrounding the legal evidentiary role of the probabilistic statements that arise in the Bayesian approach \cite{evett1987bayesian,taroni2010data}. As the introduction outlines, we apply the Bayesian approach for the scientific benefits, not as a means to establish legally admissible evidence. This approach has been successfully applied to the problem of age-at-death estimation based on skeletal remains \cite{kimmerle2008analysis,prince2008new,langley2010bayesian,brennaman2017bayesian}, and, more recently, to PMI estimation. Andersson et al. \cite{andersson2019application} take a Bayesian approach to PMI prediction based on partial body score measurements. Their model is based on indoor cases in the absence of insects. Giles et al. \cite{giles2023solving} also take a Bayesian approach but leverage a more diverse set of cases and taphonomic variables, though they represent decomposition using a TBS-like total decomposition score (TDS) \cite{gelderman2018development}. Like us, they formulate the empirical decomposition process using probabilistic models. Similarly, they formulate PMI prediction as an inverse problem: a problem where one observes the effect and needs to infer the cause. Our primary departure from Giles et al. \cite{giles2023solving} is that we directly model the disaggregated decomposition characteristics present in the geoFOR ontology, such as skin slippage, marbling of the skin, and desiccation. In addition, the geoFOR dataset includes a more diverse set of cases from research facilities and medicolegal death investigations. Lastly, we also demonstrate the use of the Bayesian approach for the design of new taphonomic experiments.

\section{Material and Methods}

\begin{table}[t!]
  \centering
  \resizebox{\textwidth}{!}{
    \begin{tabular}{l l} 
      \multicolumn{2}{ c }{\textbf{Covariates}}                                              \\\hline
      \noalign{\vskip 3px}
      \multicolumn{2}{ c }{\textit{Insect/scavenger activity}}                               \\
      Fly eggs (10\%)                                & Larva (31\%)                          \\
      Pupae (7.4\%)                                  & Adult flies (17\%)                    \\
      Ants (3.3\%)                                   & Beetles (8.3\%)                       \\
      Other insect activity (8.1\%)                  & Rodent activity (0.57\%)              \\
      Carnivore activity (0.83\%)                    & Vultures (0.13\%)                     \\
      Other scavenger activity (0.087\%)             &                                       \\
      \noalign{\vskip 5px}
      \multicolumn{2}{ c }{\textit{Other covariates}}                                        \\
      Deposition site type                           & Age                                   \\
      Body size estimation                           & Presence of clothing                  \\
      Evidence of trauma                             & Sex                                   \\
      Hanging (0.70\%)                                                                       \\
      \noalign{\vskip 8px}
      \multicolumn{2}{ c }{\textbf{Decomposition variables}}                                 \\\hline
      \noalign{\vskip 3px}
      Fresh - livor mortis absent (1.6\%)            & Livor mortis unfixed (8.2\%)          \\
      Livor mortis fixed (31\%)                      & Fresh - rigor mortis absent (4.1\%)   \\
      Rigor mortis partial (6.9\%)                   & Rigor mortis full (8.2\%)             \\
      Body intact but rigor mortis has passed (35\%) & Corneal clouding (14\%)               \\
      Drying of fingertips, lips and/or nose (32\%)  & Greening of the abdomen (27\%)        \\
      Skin slippage (48\%)                           & Skin discoloration (54\%)             \\
      Marbling (33\%)                                & Bloat (29\%)                          \\
      Purging (30\%)                                 & Adipocere (2.2\%)                     \\
      Abdominal caving (5.6\%)                       & Liquid decomposition (13\%)           \\
      Desiccation (31\%)                             & Exposed bone with moist tissue (18\%) \\
      Exposed bone with desiccated tissue (14\%)     & Weathered bone (1.9\%)                \\
      Bone with grease (3.6\%)                       & Dry bone (3.7\%)                      \\
    \end{tabular}
  }
  \caption{Variables in the geoFOR dataset. The covariates are a mixture of binary and categorical variables. All of the decomposition characteristics are binary. In the case of binary variables, the percentages represent the share of cases with the trait.}
  \label{tab:geofor_vars}
\end{table}

\begin{table}[h!]
  \centering
  \resizebox{\textwidth}{!}{
    \begin{tabular}{ l  p{9 cm} }
      \textbf{Covariate name}       & \textbf{Levels}                                                    \\\hline
      \noalign{\vskip 3px}
      \textbf{Deposition site type} & *Surface (38\%), Shallow Burial (0.22\%),
      Water (4.5\%), Structure (54\%), Vehicle (2.1\%), Unknown (1.2\%)                                  \\
      \noalign{\vskip 4px}
      \textbf{Body size estimation} & Obese (18\%), Emaciated (8.3\%), *Moderate (67\%), Unknown (6.9\%) \\
      \noalign{\vskip 4px}
      \textbf{Evidence of trauma}   & Unknown (5.1\%), *Absent (65\%), Present (30\%)                    \\
      \noalign{\vskip 4px}
      \textbf{Presence of clothing} & *Fully Clothed (23\%), Partially Clothed (24\%),
      Unclothed (45\%), Unknown (8.2\%)                                                                  \\
      \noalign{\vskip 4px}
      \textbf{Age}                  & Infant (0.74\%), Child (0.31\%), *Adult (99\%)                     \\
      \noalign{\vskip 4px}
      \textbf{Sex}                  & *Male (66\%), Female (34\%), Unknown (0.21\%)
    \end{tabular}
  }
  \caption{Levels for each of the categorical variables in geoFOR. Stars denote the reference levels. Our models measure the effect of each level relative to the reference. The percentages represent the share of cases at each level.}
  \label{tab:categorical_levels}
\end{table}

\subsection{Taphonomic case data}

As discussed above, the geoFOR project uses crowdsourcing to collect standardized taphonomic case records. We use this data to train and evaluate our models describing the appearance of decomposition characteristics. At the time of our analysis, the dataset included case records from 2,529 deceased individuals from a mixture of medicolegal (1,779) and research facility (750) sources\footnote{At the time of submission, the dataset had grown to roughly 3,200 cases.}. Each case record consists of an estimate of the date of death, 18 covariates\footnote{Throughout this work, we will use the term \textit{covariate} to refer to a case-level variable other than PMI that potentially explains some of the variation in the observed decomposition characteristics.} describing individual and environmental factors, and 24 decomposition characteristics (see Table \ref{tab:geofor_vars}). When building our models, we leverage the same set of cases used in Weisensee et al.~\cite{weisensee2024geofor}. We refer readers to that work for a complete description of the geoFOR dataset.

All \textit{Insect/scavenger activity} covariates and all decomposition variables listed in Table \ref{tab:geofor_vars} have binary outcomes indicating the absence or presence of the feature. Except for ``Hanging", the \textit{Other covariates} are categorical variables with more than two possible outcomes. Table \ref{tab:categorical_levels} lists the possible levels for each categorical variable. For the categorical variables, respondents may answer ``Unknown" or skip a question entirely, leading to a missing value. We group these cases into a single ``Unknown" category. The one exception to this rule is the ``Age" variable. In this case, missing and ``Unknown" values are imputed as ``adult," as over $99\%$ of cases in the geoFOR dataset are in the adult category.

\begin{figure}[h!]
  \centering
  \includegraphics[width=0.6\textwidth]{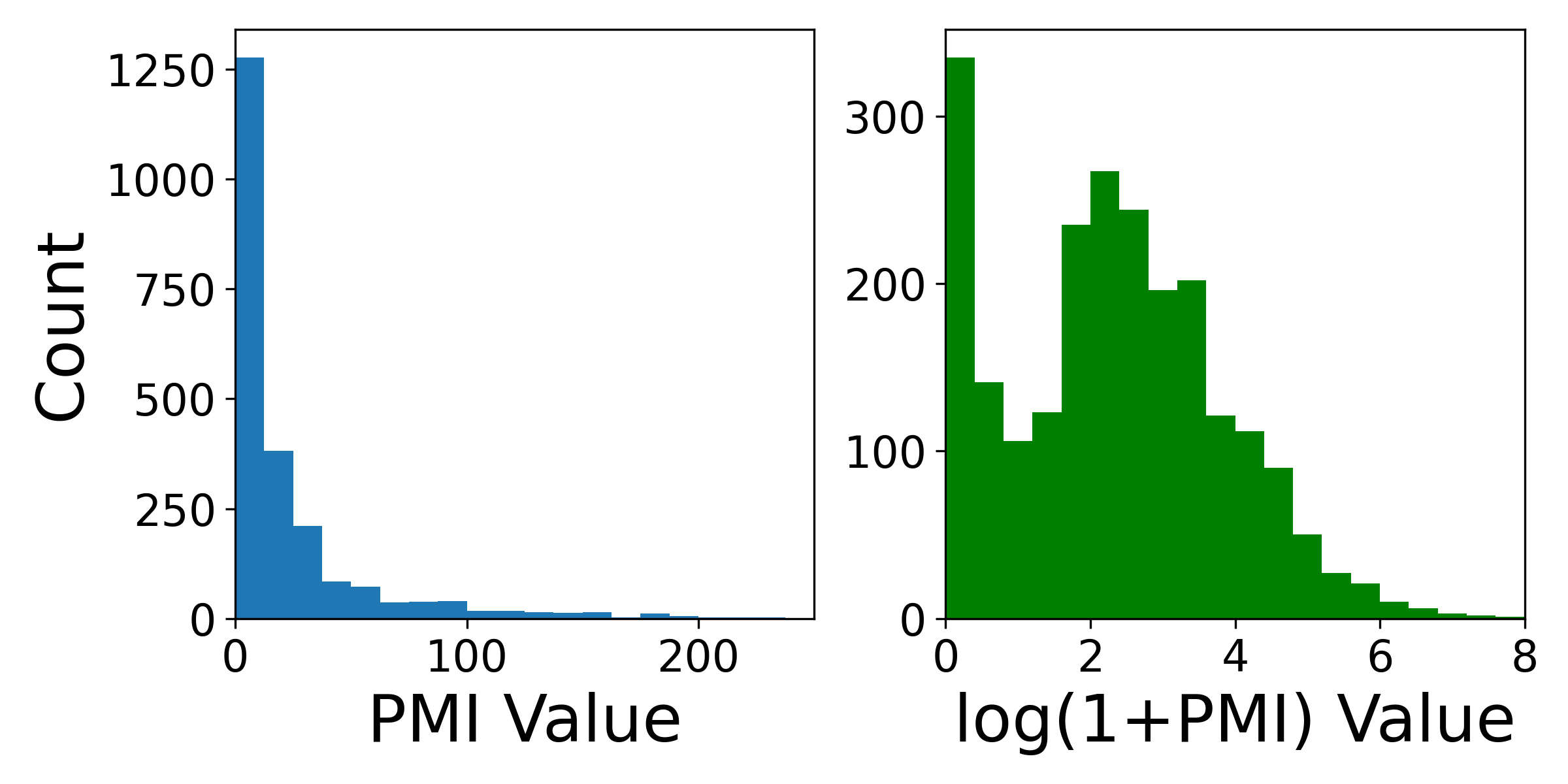}
  \caption{Histogram of PMI values in the geoFOR dataset along with log-transformed PMI values discussed in Section \ref{sec:likelihood}.}
  \label{fig:pmi_hist}
\end{figure}

When modeling decomposition, it is crucial to know the PMI. Each case includes the discovery date and the approximate or exact date of death or the date when the individual was last known to be alive. We use these dates to estimate the PMI. We can compute the PMI very accurately for research facility cases (N=752) or medicolegal cases where the date of death is known exactly (N=180). We expect our estimates to have some error for the remaining cases (N=1,597). We disregard this error in the analysis reported in this paper. See Supplemental Section \ref{app-sec:pmi-calc} for more information about our methods for PMI calculation. Figure \ref{fig:pmi_hist} shows the distribution of PMI values.

We will use the symbol $\mathcal{D}$ to refer to the geoFOR case dataset. This dataset contains triples of the form $(\supp{t}{n},\,\supp{\mathcal{X}}{n},\,\supp{\mathcal{Y}}{n})$ where $n$ selects a specific case among the $N$ cases in geoFOR. For case $n$, $\supp{t}{n}$ is the observed PMI. $\supp{\mathcal{X}}{n}$ and $\supp{\mathcal{Y}}{n}$ are the sets of covariate and decomposition values for case $n$. We refer to covariate $c$ as $\supp{x}{n}_c$ and decomposition characteristic $d$ as $\supp{y}{n}_d$. Refer to Supplemental Table \ref{app-tab:notation} for a glossary of our mathematical notation.

\subsection{Modeling Decomposition}

\subsubsection{Representing the Mechanisms of Decomposition}\label{sec:likelihood}
As represented in Figure \ref{fig:graphical_abstract}, we design a model for decomposition where the probability of observing a decomposition characteristic is a function of the PMI and the covariates. But why do we model decomposition and not PMI directly? Through a mechanistic lens, the passage of time and the covariates cause the observed state of decomposition and not the other way around. For example, we may hypothesize that clothing influences the onset of desiccation, but whether a corpse is desiccated has no causal influence on whether or not the individual was clothed. Thus, a model for decomposition represents the physical mechanisms at work more directly than a model for PMI. Concretely, the inferred model parameters that relate (for example) ``Presence of clothing" to the onset of ``Desiccation'' give us generalizable, interpretable insight into the influence of clothing on the decomposition process. We believe this sort of insight is essential to the science of forensic taphonomy.

On the other hand, a model that directly predicts PMI as a function of the covariates and state of decomposition must do so by approximating the inverse of the underlying mechanistic process by which decomposition occurs. Even for simple causal processes, the inverse model can be highly complex. This complexity makes it impractical to extract the underlying mechanisms from the fit model. Such a model may have excellent PMI predictive performance but shed no light on scientific questions about decomposition. We do not deny the utility of such a model; accurate PMI prediction has much forensic value. Instead, we point out the limits of a PMI prediction approach for scientific advancement in forensic taphonomy.

In specifying our decomposition model, we do not attempt to represent the many physicochemical mechanisms that relate the covariates and PMI to the observed state of decomposition. Instead, we build simple empirical abstractions into our model that represent high-level phenomena. For instance, some versions of our model have a parameter that represents how much more likely it is for a corpse with a given PMI to be desiccated if they are unclothed. This parameter bundles together many underlying physiochemical mechanisms, preventing us from drawing any conclusions about those processes. However, our approach has the advantage that the model can be fit based on the easily observed characteristics present in the geoFOR dataset. Correspondingly, the model effects (such as the effect of clothing on desiccation) can be easily understood and applied by forensic investigators. High-level findings from models such as ours can guide further research into the lower-level mechanisms of decomposition.

Having decided to model decomposition and not PMI directly, we must still decide \textit{how} to represent the relationships between the causes (PMI and covariates) and the effects (the decomposition characteristics). We propose a simple model structure where each relevant covariate contributes linearly to the rate at which decomposition characteristics become more likely with increasing PMI. To represent this model mathematically, we define a total case-specific rate coefficient $\supp{B}{n}_d$ such that the estimated log-odds of observing decomposition characteristic $d$ for case $n$ becomes
\begin{equation} \label{eq:liklihood}
  \mathrm{logit}\, P\left(\supp{y}{n}_d=1|\supp{t}{n}, \supp{B}{n}_d \right) =
  \gamma_d + \log(1+\supp{t}{n}) \supp{B}{n}_d
\end{equation}
where $\mathrm{logit}\,p = \log(p/(1-p))$, and $\gamma_d$ is the log-odds of observing $d$ on the date of death. $\supp{B}{n}_d$ is the cumulative effect of all of the covariates observed for case $n$:
\begin{equation}\label{eq:B}
  \supp{B}{n}_d = \beta_{d0} + \sum_{c=1}^{C_d} \beta_{dc\ell}|_{\ell=\supp{x}{n}_c}
\end{equation}
where the sum is over the set of covariates that influence decomposition variable $d$.
$\beta_{dc\ell}$ is the contribution to $\supp{B}{n}_d$ from observing that case $n$ had level $\ell$ on covariate $c$. For example, one such contribution might be the effect on ``Desiccation" from observing that a case had level ``Unclothed" for the ``Presence of clothing" covariate. The summation adds all of these contributions together. The first term in Equation \ref{eq:B}, $\beta_{d0}$, absorbs the effects of the reference levels from each covariate. Refer to Supplementary Table \ref{app-tab:notation} for a glossary of our mathematical notation.

The log transformation of $\supp{t}{n}$ in Eq.~\eqref{eq:liklihood} implies that squaring the PMI duration results in a near doubling of the size of an effect, leading to diminishing changes at large PMI values. Though there is no in-principle reason why this mathematical formulation is the true description of how decomposition unfolds through time, it works well in practice. Similarly, we could modify the time and covariate structure in equations \eqref{eq:liklihood} and \eqref{eq:B} in many ways to represent various decomposition mechanisms more precisely. For example, some decomposition variables, such as rigor mortis, will appear early during decomposition and then later disappear. Our model does not adequately capture this structure. We do not claim that our formulation is the \textit{right} one. We propose this particular structure because it achieves a good balance between the desire to model the observed data accurately and the desire to facilitate the interpretation of the fitted model parameters in terms of the mechanisms of decompositions. In addition, we hope this simple approach will provide a valuable baseline for future extensions of this work.

\subsubsection{Choosing Interactions}\label{sec:model_variants}

We designed our model to represent mechanistic relationships between the covariates and the decomposition characteristics. However, many covariates have no plausible relationship to the decomposition characteristics. For instance, clothing should not influence the onset of corneal clouding. For this reason, we have created three model variants called ``empty," ``strict," and ``full" models. The ``empty" model removes all covariates and only considers the effect of PMI. The ``strict" model only includes effects for covariate-decomposition pairs if the covariate was very likely to influence the decomposition characteristic, according to two experienced forensic anthropologists on our team. The ``full" allows all covariates to influence all decomposition characteristics. Table \ref{tab:strict} records our selections for which covariates influence which decomposition characteristics in the ``strict'' model.

\begin{table}[h!]
  \centering
  \caption{Effects included in the ``strict" model variant. Check marks indicate that the covariate is included. Age at death, Sex, Beetles, Ants, Fly Eggs, Other Insect Activity, and Other Scavenger Activity are omitted as none of these covariates are included in the ``strict" model.}
  \resizebox{\textwidth}{!}{%
    \begin{tabular}{lccccccccccc}
      \textbf{Decomposition Characteristics} & \rottwo{\textbf{PMI}}         & \rottwo{\textbf{Hanging}}  & \rottwo{\textbf{Deposition Site Type}} & \rottwo{\textbf{Body Size Estimation}} & \rottwo{\textbf{Presence of Clothing}} & \rottwo{\textbf{Evidence of Trauma}} &
      \rottwo{\textbf{Rodent activity}}      & \rottwo{\textbf{Larva}}       & \rottwo{\textbf{Vultures}} &
      \rottwo{\textbf{Carnivore Activity}}   & \rottwo{\textbf{Adult Flies}}                                                                                                                                                                                                                              \\    \hline
      \noalign{\vskip 2px}
      Desiccation                            & \OK                           & \OK                        & \OK                                    & \OK                                    & \OK                                    &                                      &     & \OK & \OK & \OK &     \\
      Skin slippage                          & \OK                           & \OK                        & \OK                                    &                                        & \OK                                    &                                      &     &     &     &     &     \\
      Exposed bone moist tissue              & \OK                           & \OK                        & \OK                                    & \OK                                    & \OK                                    & \OK                                  & \OK & \OK & \OK & \OK &     \\
      Exposed bone desiccated tissue         & \OK                           & \OK                        & \OK                                    & \OK                                    & \OK                                    & \OK                                  & \OK & \OK & \OK & \OK &     \\
      Bloat                                  & \OK                           & \OK                        & \OK                                    &                                        &                                        &                                      &     & \OK &     &     &     \\
      Purging                                & \OK                           & \OK                        & \OK                                    & \OK                                    &                                        &                                      &     &     &     &     &     \\
      Bone with grease                       & \OK                           & \OK                        & \OK                                    & \OK                                    & \OK                                    & \OK                                  & \OK & \OK & \OK & \OK &     \\
      Dry bone                               & \OK                           & \OK                        & \OK                                    & \OK                                    & \OK                                    & \OK                                  & \OK &     & \OK & \OK &     \\
      Adipocere                              & \OK                           & \OK                        & \OK                                    &                                        &                                        &                                      &     & \OK &     &     &     \\
      Abdominal caving                       & \OK                           & \OK                        & \OK                                    &                                        &                                        &                                      &     & \OK &     &     &     \\
      Weathered bone                         & \OK                           & \OK                        & \OK                                    &                                        & \OK                                    &                                      & \OK & \OK & \OK & \OK &     \\
      Body intact, rigor mortis passed       & \OK                           & \OK                        & \OK                                    &                                        &                                        &                                      & \OK &     & \OK & \OK & \OK \\
      Marbling                               & \OK                           & \OK                        &                                        &                                        &                                        &                                      &     &     &     &     &     \\
      Skin discoloration                     & \OK                           & \OK                        &                                        &                                        &                                        &                                      &     &     &     &     &     \\
      Greening of the abdomen                & \OK                           & \OK                        &                                        &                                        &                                        &                                      &     &     &     &     &     \\
      Drying of fingertips, lips and/or nose & \OK                           & \OK                        &                                        &                                        &                                        &                                      &     &     &     &     &     \\
      Livor mortis fixed                     & \OK                           &                            &                                        &                                        &                                        &                                      &     &     &     &     &     \\
      Fresh livor mortis absent              & \OK                           &                            &                                        &                                        &                                        &                                      &     &     &     &     &     \\
      Rigor mortis partial                   & \OK                           &                            &                                        &                                        &                                        &                                      &     &     &     &     &     \\
      Rigor mortis full                      & \OK                           &                            &                                        &                                        &                                        &                                      &     &     &     &     &     \\
      Livor mortis unfixed                   & \OK                           &                            &                                        &                                        &                                        &                                      &     &     &     &     &     \\
      Fresh rigor mortis absent              & \OK                           &                            &                                        &                                        &                                        &                                      &     &     &     &     &     \\
      Corneal clouding                       & \OK                           &                            &                                        &                                        &                                        &                                      &     &     &     &     &     \\
    \end{tabular}}
  \label{tab:strict}
\end{table}

\subsubsection{Inference procedure}\label{sec:inference}
In Equations \eqref{eq:liklihood} and \eqref{eq:B}, we introduced unknown parameters that represent the effects of various decomposition mechanisms. We refer to the set of all such parameters as $\mathcal{C}$. The goal of inference is to estimate the values of these parameters given the geoFOR dataset $\mathcal{D}$. In other words, we wish to find the distribution $p(\mathcal{C}|\mathcal{D})$. As motivated in the introduction, we follow a Bayesian inference approach. Bayesian inference finds the values of $\mathcal{C}$ that \textit{explain the observed data} while also \textit{being consistent with our prior beliefs} about the unknown coefficients. We can express the notion of explaining the data as maximizing the likelihood of the data given the coefficients: $p(\mathcal{D}|\mathcal{C})$. Likewise, we express the notion of being consistent with our prior beliefs as maximizing the likelihood under the prior $p(\mathcal{C})$. Bayes rule then tells us that
\begin{equation}\label{eq:bayes}
  p(\mathcal{C}|\mathcal{D}) \propto p(\mathcal{D}|\mathcal{C})p(\mathcal{C}).
\end{equation}
The proportionality constant does not depend on $\mathcal{C}$ and can be ignored. Rather than compute $p(\mathcal{C}|\mathcal{D})$ directly, we use well-established Markov chain Monte Carlo (MCMC) algorithms to sample $N_\mathcal{C}$ values from this distribution. We can then use these samples to describe effect sizes and make inferences about other quantities of interest. See Supplemental Section \ref{app-sec:priors} for our specification of $p(\mathcal{C})$.

\subsubsection{Evaluation methodology}

We evaluate our model by partitioning $\mathcal{D}$ into train and test parts, fitting the model on the train part and testing on the test part. For robustness, we perform this evaluation in a cross-validation framework. We randomly split the data into $k$ folds. We then iteratively treat each of the $k$ folds as the test part and the remaining $k-1$ folds as the train part. This procedure gives us $k$ semi-independent evaluations, which we aggregate to estimate model performance robustly. We use $k=5$. We apply this evaluation process to the ``empty", ``strict", and ``full" model variants described in Section \ref{sec:model_variants}.

Given the PMI and covariates, our model estimates the 24 decomposition characteristics in geoFOR. We evaluate our model's success at this task using the ROC Area Under the Curve (ROC AUC) \cite{bradley1997use}. A perfect model has an ROC AUC of 1, and an ROC AUC of 0.5 is no better than randomly guessing the decomposition variables. We estimate this area using the $k=5$ test parts and average to get a final estimate. We average the ROC AUC scores over the 24 classes to get a single model performance number. We also present averaged ROC curves for each model variant.

\subsection{PMI inference}\label{sec:pmi-inference}
The posterior distribution $p(\mathcal{C}|\mathcal{D})$, defined in Eq.~\ref{eq:bayes}, can be used to make inferences about PMI for new cases. Let $(\mathcal{X}, \mathcal{Y})$ be the record for a new case that was not used when inferring $p(\mathcal{C}|\mathcal{D})$. This record does not contain the PMI $t$. This inference step aims to estimate $t$ given the observed covariates $\mathcal{X}$ and decomposition variables $\mathcal{Y}$ for the case. In this estimation, we want to consider what we learned about the relationships between $t$, $\mathcal{X}$, and $\mathcal{Y}$ in the inference procedure described in the previous section. In other words, we compute the distribution
\begin{eqnarray}\label{eq:pmi-infer}
  p(t | \mathcal{X}, \mathcal{Y}, \mathcal{D}) &=& \int d\mathcal{C} p(\mathcal{C}|\mathcal{D}) p(t | \mathcal{X}, \mathcal{Y}, \mathcal{C}) \nonumber \\
  &\approx& \frac{1}{N_\mathcal{C}} \sum_{i=1}^{N_\mathcal{C}} p(t | \mathcal{X}, \mathcal{Y}, \supp{\mathcal{C}}{i})
\end{eqnarray}
Where $\supp{\mathcal{C}}{i}$ is a sample drawn from $p(\mathcal{C}|\mathcal{D})$ using MCMC. As shown in Supplemental Section \ref{app-sec:pmi}, $p(t | \mathcal{X}, \mathcal{Y}, \supp{\mathcal{C}}{i})$ can be calculated explicitly from Eq.~\eqref{eq:liklihood} and a prior distribution $p(t)$ through the application of Bayes rule.

We measure our model's ability to predict PMI using the $R^2$ metric. Following the original geoFOR publication \cite{weisensee2024geofor}, we compute this metric in the log-transformed PMI space: $\log(1+\supp{T}{n})$. We estimate $R^2$ for each of the $k=5$ folds and average to get a robust measure of our model's ability to predict PMI. We also present a plot of predicted vs. actual PMI values and point-wise uncertainty estimates.

\subsection{Optimal experimental design}
In Optimal Experimental Design (OED) we use an existing fitted model to determine which experimental designs will likely yield the most information about specific decomposition mechanisms. Consider, for example, a research team studying the impact of age at death on the onset of desiccation. The team can only gain access to a small number of cadavers and must decide how many bodies should be placed in each age condition and how long the bodies should be observed. We refer to this set of decisions as the {\it experimental design}, $\delta$. OED selects among possible designs by maximizing the expected reduction in uncertainty for the model variables that represent the relationships under study, in this case, the variables relating to age at death and desiccation. We refer to this set of variables of interest as the {\it target variables} and denote them as $\Theta$.

We quantify the expected reduction in uncertainty using the Expected Information Gain (EIG), which is the expected magnitude of the entropy decrease for the distribution of $\Theta$ after the experiment:
\begin{equation}
  \mathrm{EIG}(\delta) = E_{y\sim p(y|\delta)}\left[H(\Theta) - H(\Theta|y, \delta)\right].
\end{equation}
$H(\Theta)$ is the entropy of the target variables before running the experiment, and $H(\Theta | y, \delta)$ is the entropy after running the experiment and obtaining outcome $y$. Since we do not know the outcome, we must take the expectation over outcomes given the design. The optimal design minimizes the information gain:
\begin{equation}
  \delta^* = \mathrm{argmin}_\delta \, \mathrm{EIG}(\delta).
\end{equation}
On average, this design will lead to the largest uncertainty reduction about $\Theta$. In practical terms, this experimental design will teach us more about the mechanism of interest than any other design we could execute. See Appendix \ref{app-sec:eig} for complete mathematical details.

\section{Results and discussion}

We present our results from fitting the decomposition model to the geoFOR dataset. We first evaluate the model's performance when predicting decomposition and PMI. We then inspect the fitted model to see what it can teach us about decomposition mechanisms. Finally, we present a case study on applying optimal experimental design to the geoFOR dataset.

\subsection{Decomposition and PMI Prediction}

Table \ref{tab:results-pred} shows the cross-validated performance of the three model variants. Unsurprisingly, the model's decomposition predictions improve as we add more covariates, as indicated by larger ROC AUC scores for the ``Strict'' and ``Full'' models. However, even the ``Empty" model does significantly better than a random baseline (ROC AUC = 0.5), suggesting that PMI alone strongly predicts decomposition characteristics. 

\begin{table}[h!]
  \centering
  \begin{tabular}{ll*{6}{c}r}
    \centering
    \textbf{Prediction task} & \textbf{Metric} & \textbf{Empty}  & \textbf{Strict} & \textbf{Full}   \\
    \hline
    \noalign{\vskip 2px}
    Decomposition            & ROC AUC         & 0.74 $\pm$ 0.01 & 0.79 $\pm$ 0.01 & 0.85 $\pm$ 0.01 \\
    PMI                      & $R^2$           & 0.71 $\pm$ 0.03 & 0.65 $\pm$ 0.04 & 0.61 $\pm$ 0.05 \\
  \end{tabular}
  \caption{Cross-validated performance of the three model variants. The $\pm$ values represent the 95\% confidence interval estimated using 5-fold cross validation. See Supplemental Section \ref{app-sec:roc_curves}  for ROC curves for each model variant.}
  \label{tab:results-pred}
\end{table}

Surprisingly, adding more covariates harms the model's ability to predict PMI, as indicated by the lower $R^2$ score for the ``Strict'' and ``Full'' models. We hypothesize that the additional covariates in the model are masking the influence of PMI on the decomposition characteristics. This masking could occur if PMI correlates with the covariates, allowing for strong decomposition prediction without correctly modeling the direct effect of PMI. We plan to investigate this hypothesis in future work. The ``Empty" model's $R^2$ of 0.71 $\pm$ 0.03 is outperformed by the result reported in Weisensee et al.~\cite{weisensee2024geofor} of 0.814 $\pm$ 0.004. Nevertheless, the proximity of these $R^2$ values is surprising, considering that our ``Empty'' model does not use any covariates and is not trained explicitly for PMI prediction. In contrast, the Weisensee et al. \cite{weisensee2024geofor} result used a powerful gradient-boosted tree algorithm to directly predict PMI based on the covariates discussed in this manuscript and additional weather-related covariates.

\begin{figure}[H]
  \centering
  \begin{subfigure}[b]{0.48\textwidth}
      \centering
      \includegraphics[width=\textwidth]{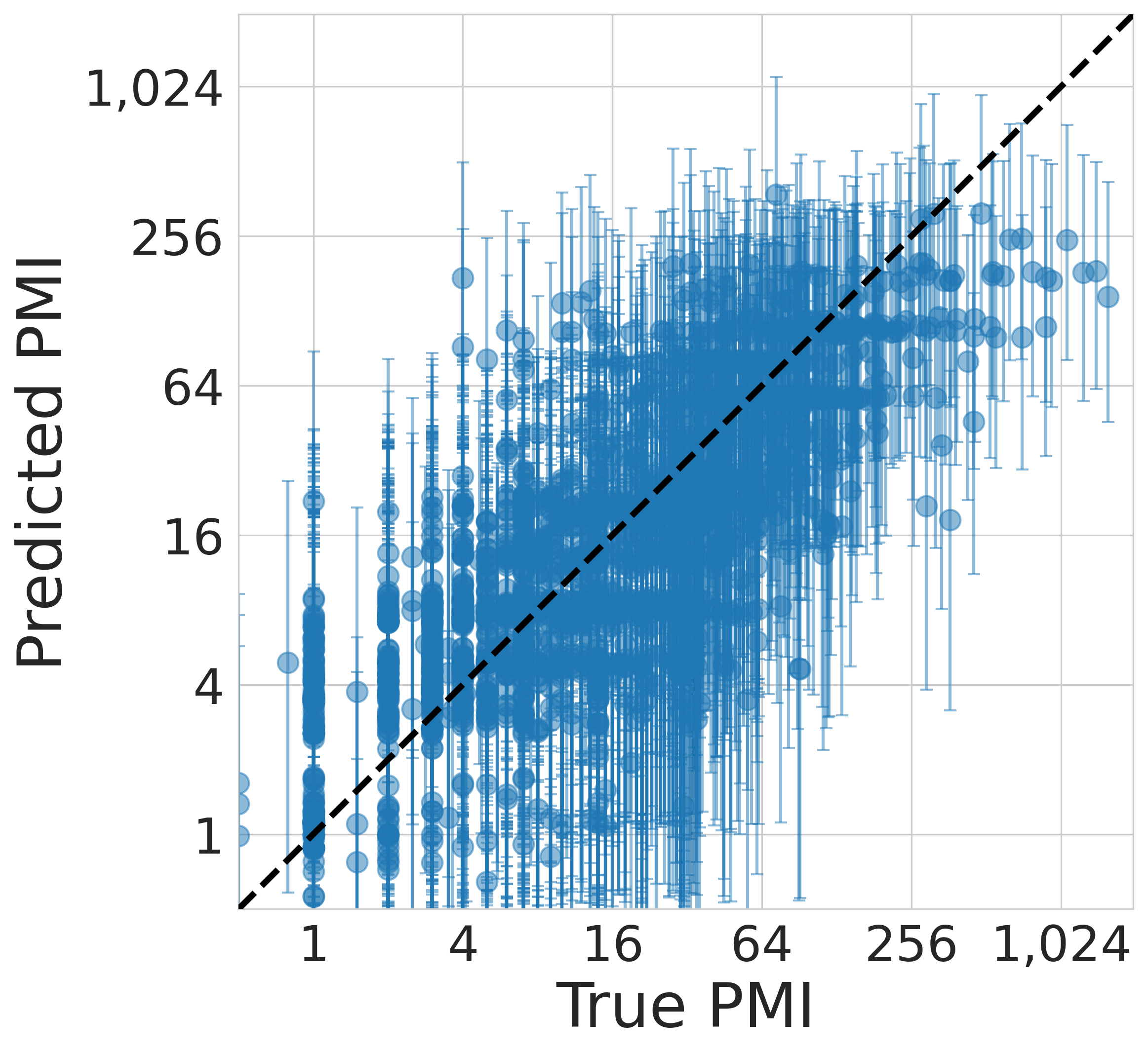}
      \caption{Predictive performance}
      \label{fig:r2_plot}
  \end{subfigure}
  \hfill
  \begin{subfigure}[b]{0.48\textwidth}
      \centering
      \includegraphics[width=\textwidth]{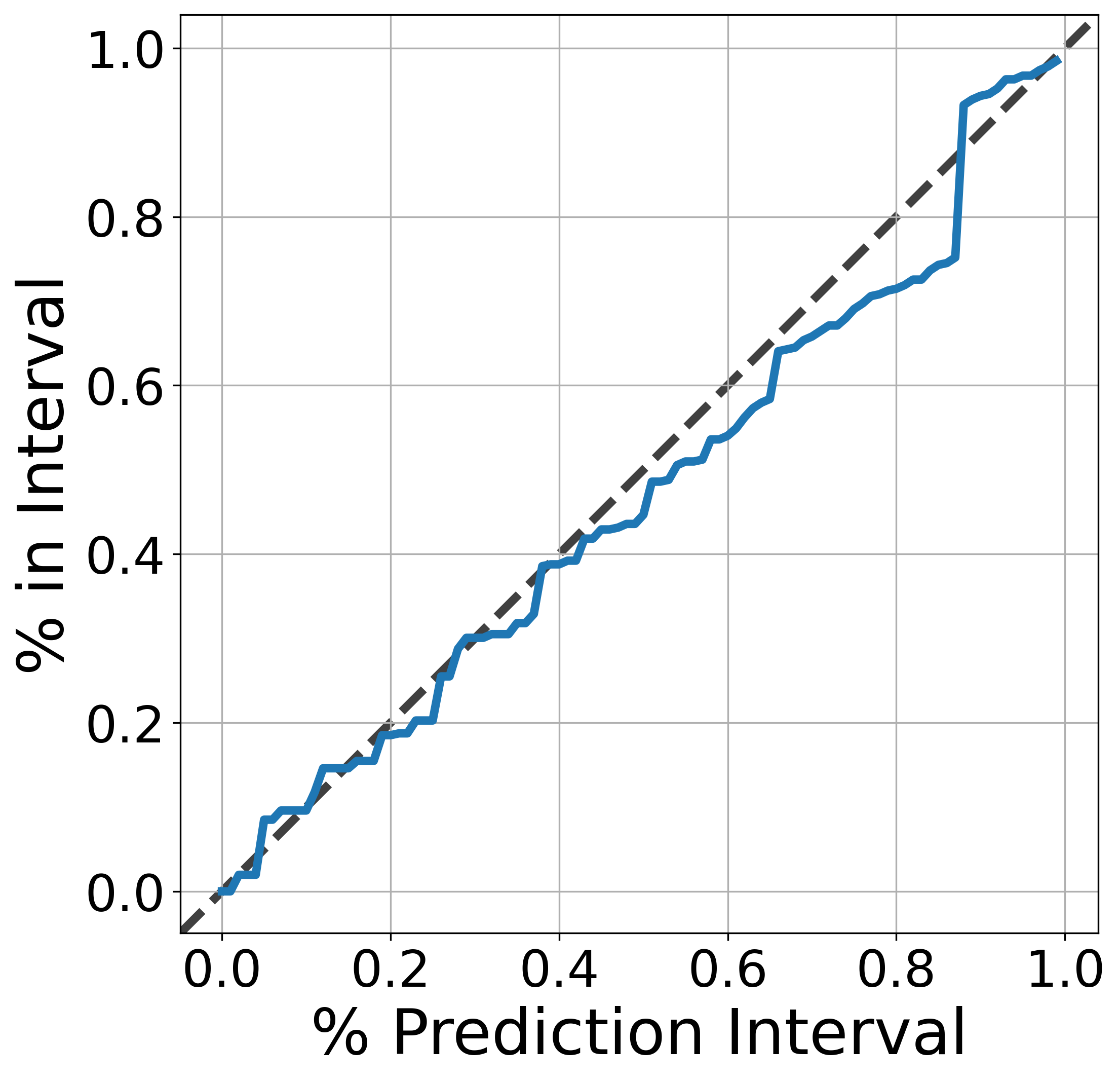}
      \caption{Calibration of uncertainty intervals}
      \label{fig:calibration}
  \end{subfigure}
  \caption{Performance of the ``Empty'' model. Figure \ref{fig:r2_plot}, left, shows the Predicted vs. True PMI values for the ``empty" model. The markers represent the mean PMI predictions, and the error bars represent the 90\% prediction interval from performing PMI posterior inference as described in Section \ref{sec:pmi-inference}. Figure \ref{fig:calibration}, right, shows the percentage of cases falling within the prediction interval as a function of the prediction interval size. A perfectly calibrated model would fall along the dashed black line.}
  \label{fig:performance_empty}
\end{figure}

Figure \ref{fig:r2_plot} shows the predicted vs. actual PMI values for the ``Empty" model. The model shows good predictive performance for PMI values up to roughly 500 PMI days. Beyond this point, the model systematically underestimates PMI. The banded structure of the plot arises due to the binary nature of the decomposition characteristics. Each horizontal band corresponds to cases with specific key decomposition characteristics. The error bars in \ref{fig:r2_plot} represent the 90\% prediction interval for each case. In other words, the model predicts a 90\% probability that the actual PMI lies within the error bars. Figure \ref{fig:calibration} tests the reliability of these prediction intervals by showing the percentage of cases that actually fall within the prediction interval as a function of the prediction interval size. For a perfect model, the percentage of cases in the interval would match the interval size and the calibration curve would fall along the dashed black line. Our model comes very close to this ideal suggesting that the prediction intervals can be relied upon for making probabilistic statements about PMI predictions.

\subsection{Decomposition mechanisms}

\subsubsection{Base decomposition rates}
\begin{figure}[H]
  \centering
  \includegraphics[width=0.9\textwidth]{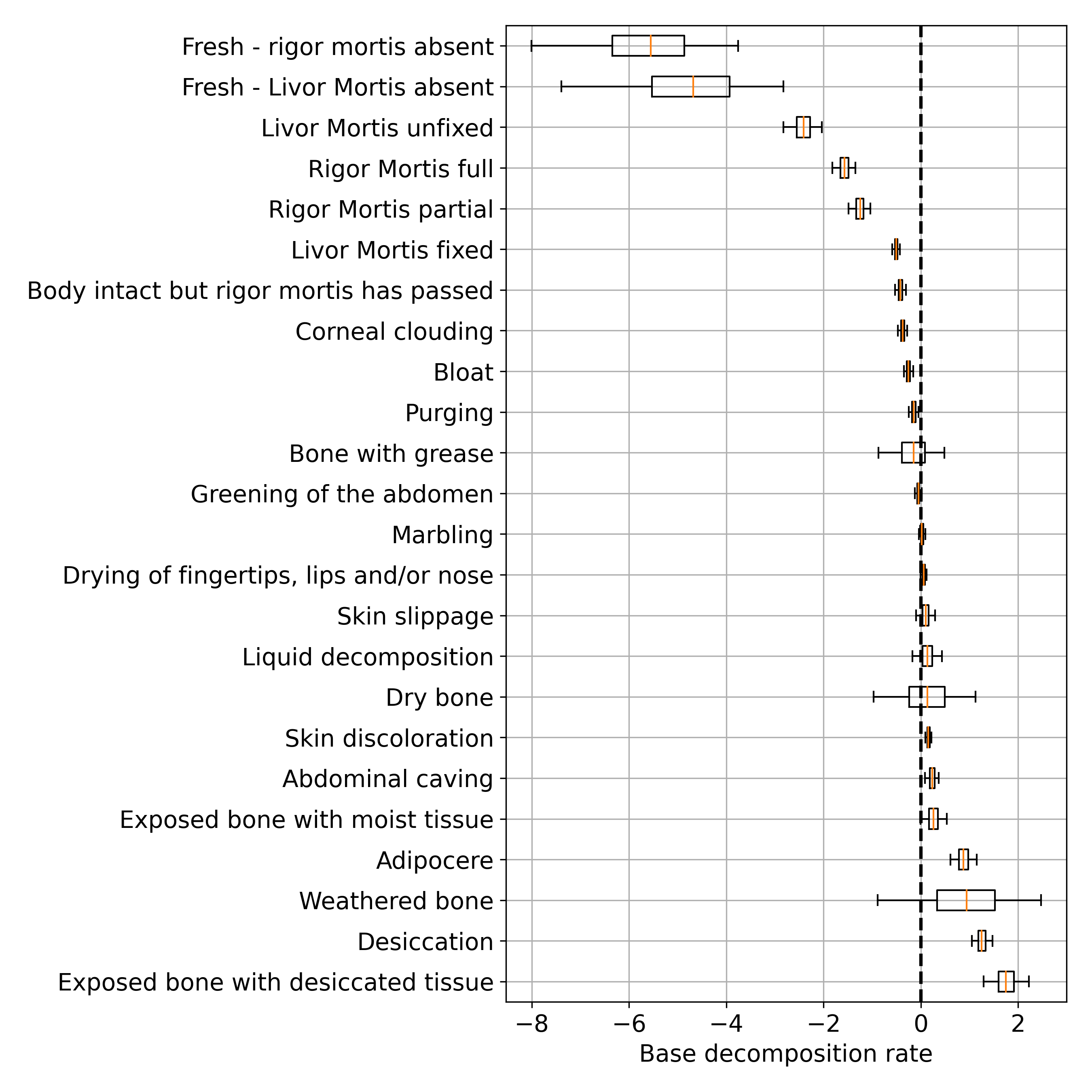}
  \caption{$\beta_{d0}$ posterior distributions for each decomposition characteristic. The distributions are ordered by the median value of $\beta_{d0}$.}
  \label{fig:boxplots}
\end{figure}

The $\beta_{d0}$ parameters in Eq.~\eqref{eq:B} quantify how the likelihood of observing decomposition variable $d$ changes with increasing PMI. Figure \ref{fig:boxplots} shows the marginal posterior distributions of $\beta_{d0}$ for each decomposition characteristic for a hypothetical case where all covariates are at their reference level. We find that the order of the median $\beta_{d0}$ values roughly corresponds to the typical temporal order in which they occur following death. Early decomposition characteristics have negative values, while late-stage characteristics have positive values. This correspondence suggests that the model captures the overall decomposition progression over time.

\subsubsection{Covariate effects}

\begin{figure}[H]
  \centering
  \includegraphics[width=0.5\textwidth]{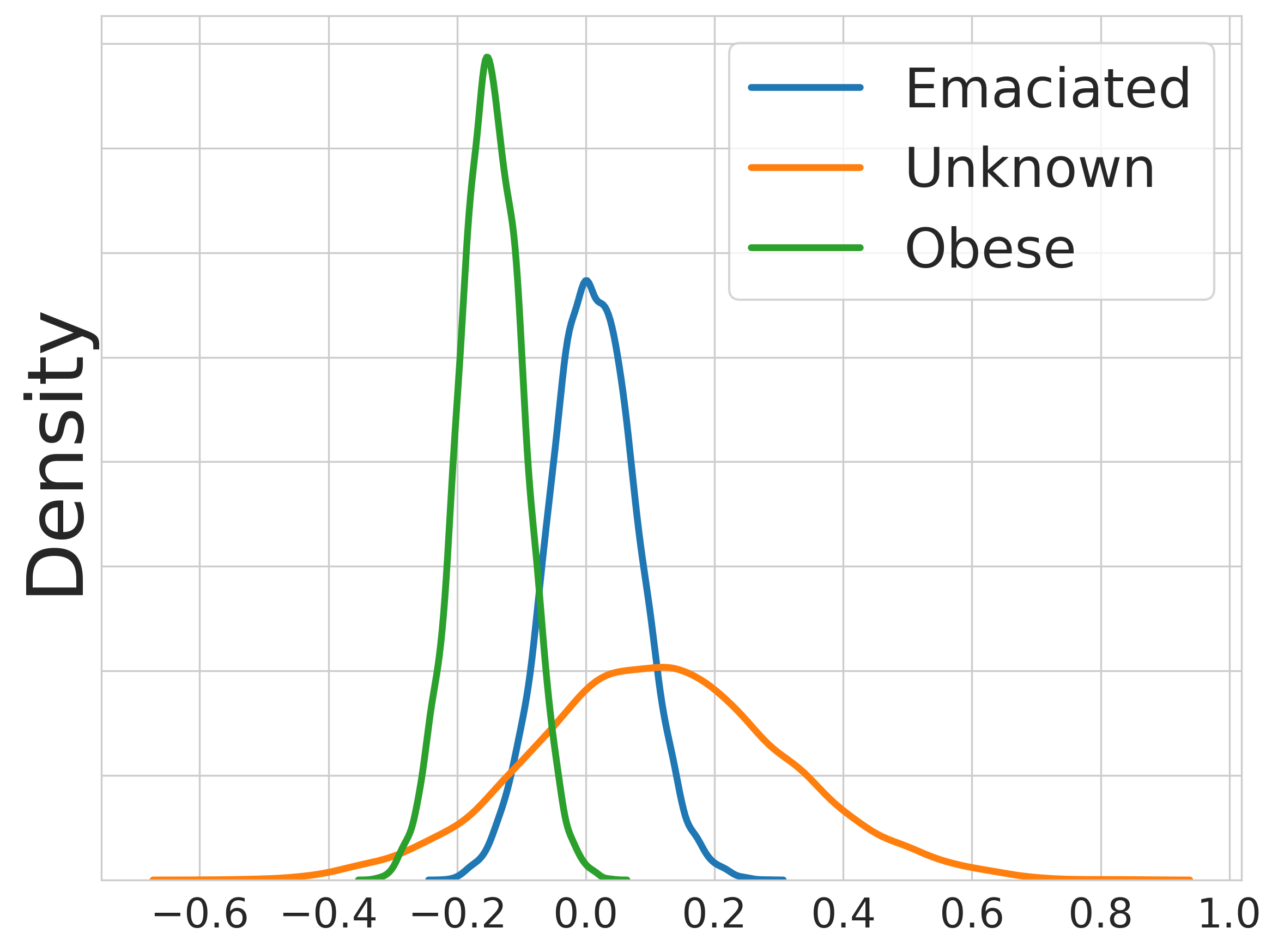}
  \caption{Probability distributions that show the relationship between the size of a body and desiccation. Negative samples correspond to desiccation being less likely in relation to normal body size. Positive samples correspond to desiccation being more likely.}
  \label{fig:body_size_dessication}
\end{figure}

The $\beta_{dc\ell}$ parameters in Eq.~\eqref{eq:B} quantify how observing level $\ell$ of covariate $c$ modifies the the rate of change in the likelihood of observing decomposition variable $d$. A positive $\beta_{dc\ell}$ indicates that level $\ell$ increases the decomposition rate. A negative value indicates a decrease in the decomposition rate. For example, Figure \ref{fig:body_size_dessication} shows the posterior distributions of $\beta_{dc\ell}$ for $d = $ ``Desiccation'' and $c = $ ``Body size''. From this, we can conclude that ``Obese'' cadavers desiccate more slowly than cadavers of moderate size. Perhaps counter-intuitively, we do not see an increase in the desiccation rate for ``Emaciated'' cadavers.

Our ``Strict" model includes 215 individual effects. We highlight several of the most interesting findings here and provide the supplementary data file called \verb|strict_model_effects.xlsx| with posterior quantiles for every effect.

\begin{figure}[h]
  \centering
  \includegraphics[width=\textwidth]{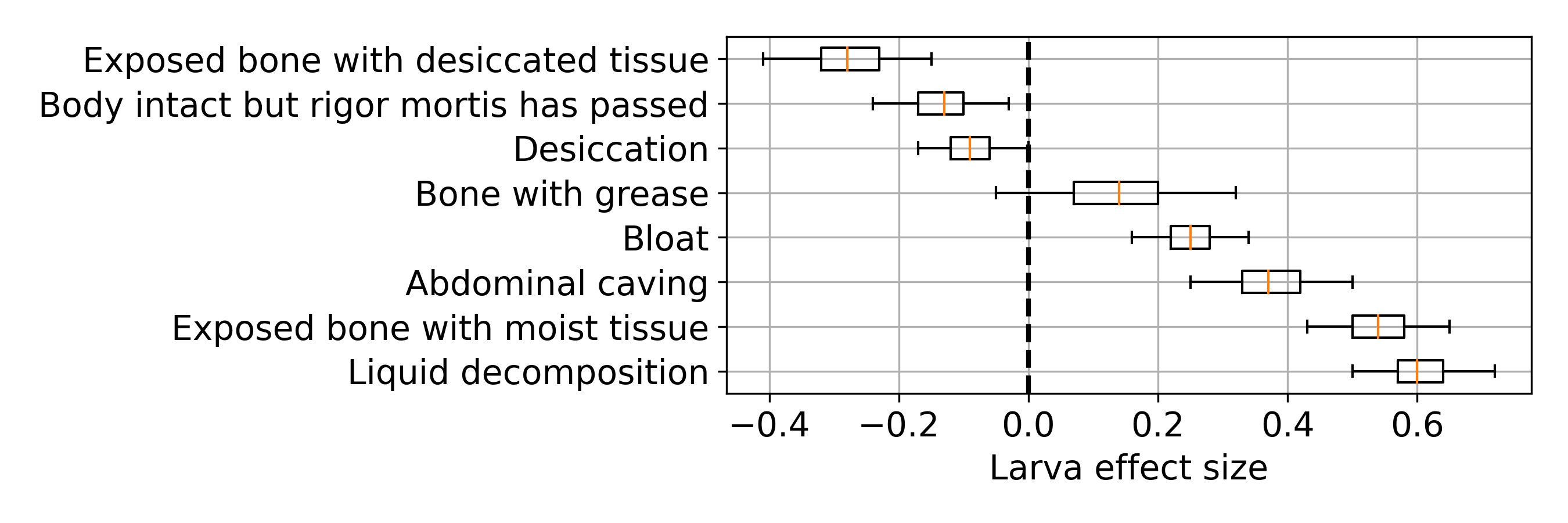}
  \caption{Effects of larva on decomposition.}
  \label{fig:effect-larva}
\end{figure}

\paragraph{Larva} As shown in Table \ref{tab:geofor_vars}, 31\% (784 out of 2,529) of geoFOR cases observed the presence of larva. This large number of cases allows the model to predict effect sizes with high confidence. Figure \ref{fig:effect-larva} shows the effect of larva on each of the eight decomposition variables that are potentially influenced by the larva in our ``Strict" model (See Table \ref{tab:strict}). Based on this plot, we conclude that larva increases decomposition rates for Bloat, Abdominal caving, Exposed bone with moist tissue, and Liquid decomposition.

The negative effect of larva on ``Exposed bone with desiccated tissue" likely results from the fact that larvae do not thrive in desiccated tissue, so desiccated tissue is absent in most cases with larvae. From this correlation, the model infers that larva has a negative effect on desiccation, which does not seem causally valid. This scenario exemplifies how the interpretation of our model effects as \textit{causes} hinges upon the causal correctness of our model specification. In this scenario, conditions in the early stages of decomposition lead to the presence of larva, which, in turn, alters the decomposition process. Our simple model formulation cannot easily capture such cyclical effects. We hope that future work can address this limitation.

\begin{figure}[h]
  \centering
  \includegraphics[width=\textwidth]{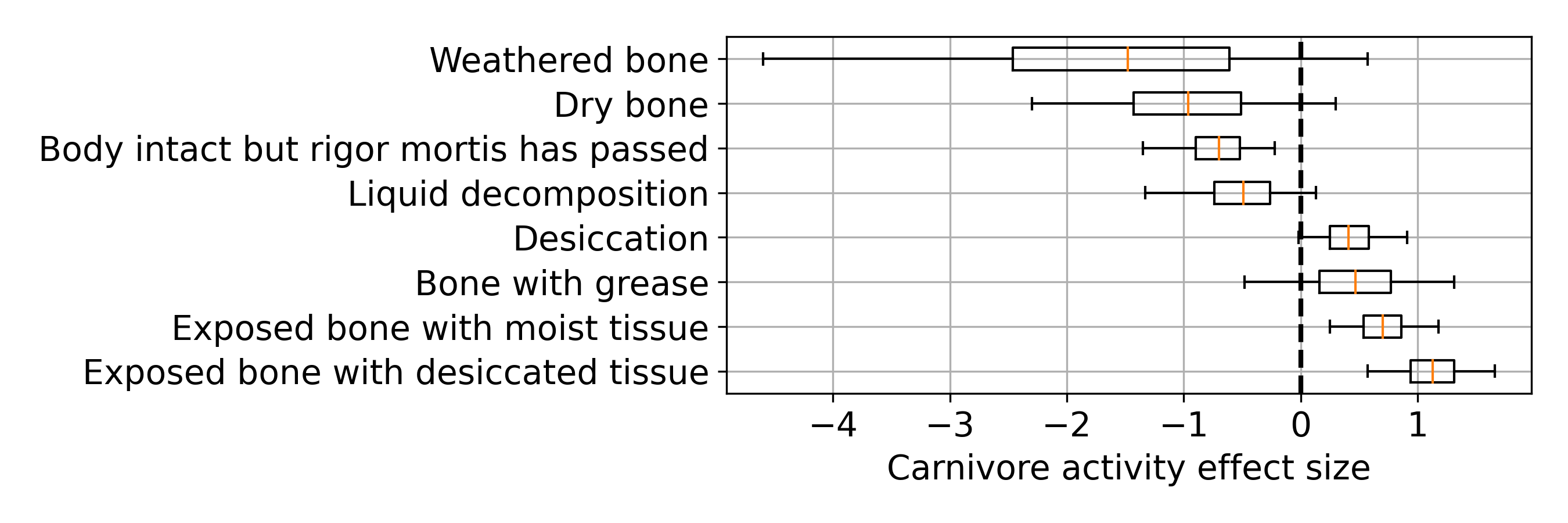}
  \caption{Effects of carnivore activity on decomposition.}
  \label{fig:effect-carnivores}
\end{figure}

\paragraph{Carnivore activity}
Unlike Larva, Carnivore activity is only present in 0.83\% (21 of 2,529) of geoFOR cases. As shown in Figure \ref{fig:effect-carnivores}, the probability intervals for the effect sizes are much broader than for larvae. Nevertheless, even with only 21 cases, we can infer that Carnivore activity leads to increased rates for Exposed bone and a decreased rate for Body intact. We also see hints of a connection between carnivore activity and desiccation, which could be explained by carnivores exposing bone and removing tissue, thus potentially contributing to an increased desiccation rate.

\begin{figure}[h]
  \centering
  \includegraphics[width=\textwidth]{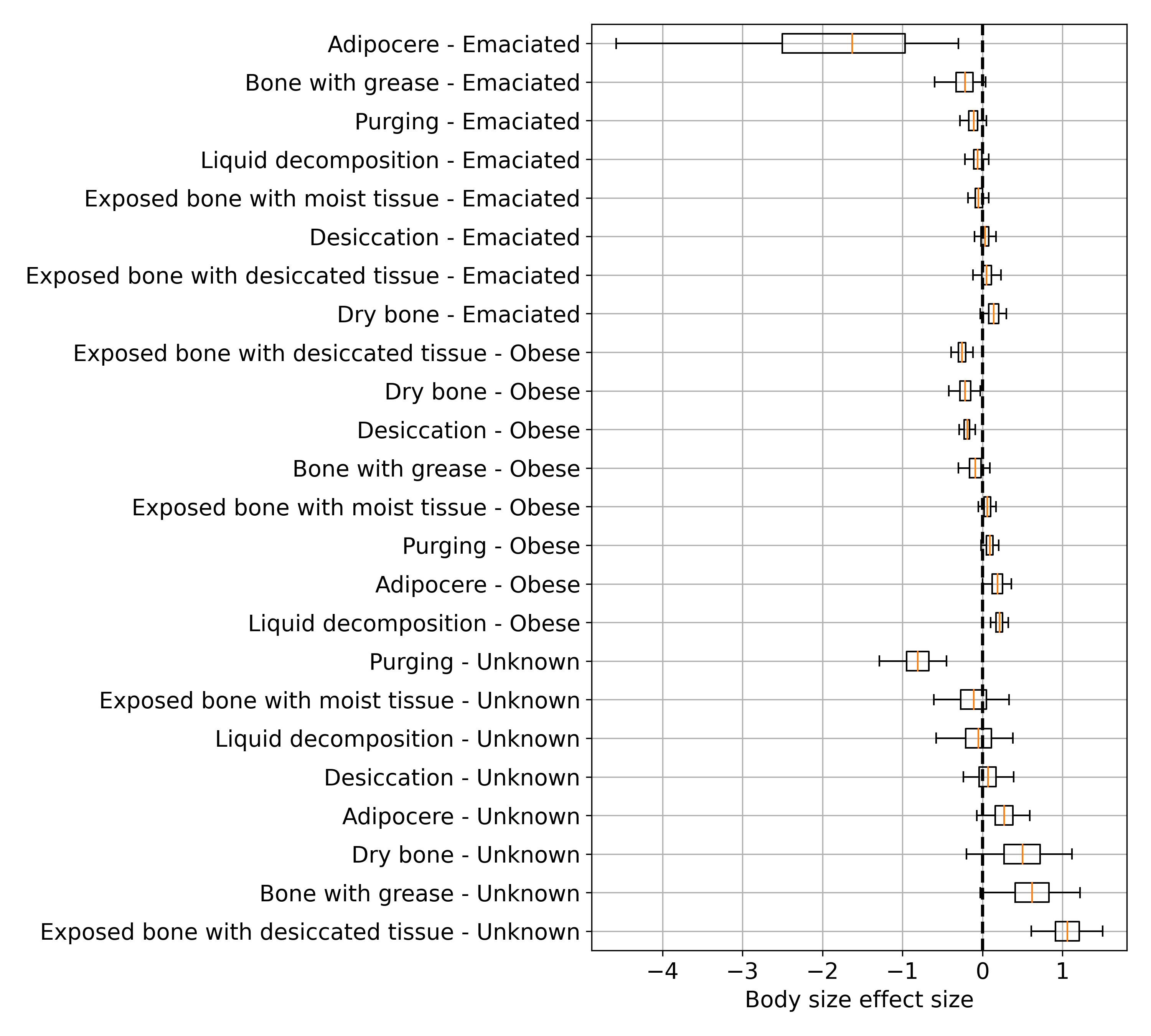}
  \caption{Effects of Body size on decomposition.}
  \label{fig:effect-bodysize}
\end{figure}

\paragraph{Body size}
Lastly, we consider the effect of ``Body size" on decomposition. As shown in Table \ref{tab:categorical_levels}, 18\% of geoFOR cases are ``Obese", 8.3\% are ``Emaciated", 6.9\% are ``Unknown", and the remaining 67\% are at the reference level (``Moderate''). Figure \ref{fig:effect-bodysize} shows the effects relative to the reference level on the six affected decomposition variables. Each level has nuanced, significant effects on the decomposition process. Many of the effects are too uncertain to distinguish from zero confidently.

We note that the ``Unknown'' level effects should not be interpreted through a causal lens. We hypothesize that the unknowns usually arise when bodies are found in the late stages of decomposition. As a result, the presence of ``Unknown'' will be correlated with late-stage variables and anti-correlated with early-stage variables. This correlation will lead to significant effects due to the limitations of observation and not due to causal mechanisms.

\subsection{Optimal Experimental Design}
We present three case studies showing how to use the Expected Information Gain (EIG) formalism to inform the design of future experiments. We imagine a research team designing an experimental protocol to understand particular decomposition mechanisms better. For demonstration, we consider the following three effects:
\begin{enumerate}
  \item Effect of ``Vultures" on ``Bone with grease''
  \item The effect of ``Age at Death" being ``Infant'' on ``Desiccation''
  \item The effect of ``Body size" being ``Emaciated'' on ``Desiccation''
\end{enumerate}
The goal is to decide how to structure the protocol to best use the limited resources (time, number of cadavers, etc.) available to the research team. Using EIG, we can estimate the reduction in uncertainty about these three effects given different experimental designs. We emphasize that these are simply examples. The constraints and goals of a particular research project will determine which effects are most important to study.

\begin{figure}[h]
  \centering
  \includegraphics[width=\textwidth]{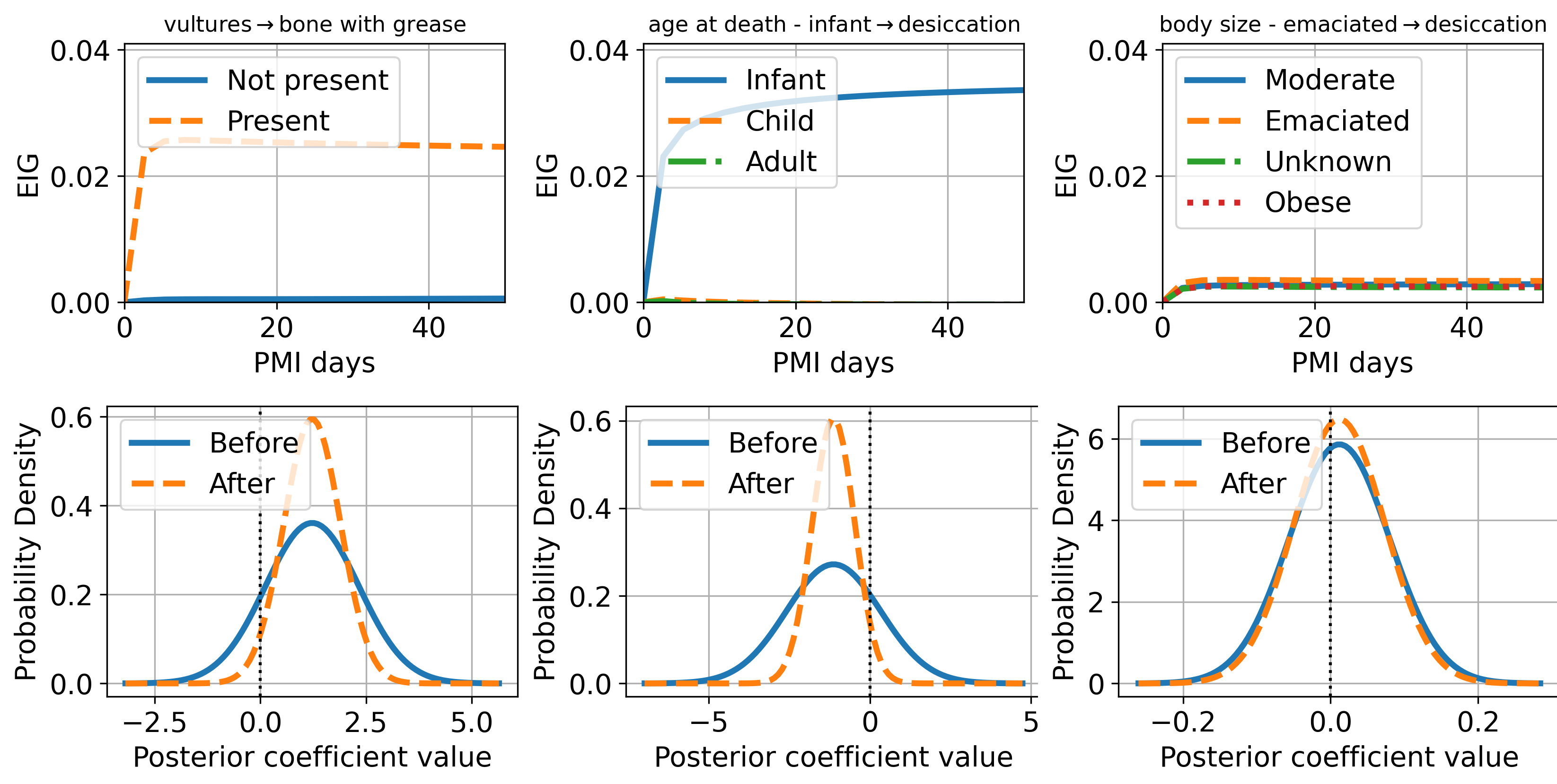}
  \caption{Optimal experimental design results. The top row shows the EIG per cadaver as a function of the number of PMI days observed. The bottom row shows the posterior distribution for the effect of interest before and after the hypothetical experiment with 30 cadavers. In the top row, the legend values refer to the chosen experimental design. In the top-left plot, ``Present'' indicates that vulture scavenging was present in the hypothetical experiment. In the bottom row, "After" refers to the expected state of knowledge after the experiment.}
  \label{fig:oed_example}
\end{figure}

For our demonstration, we set the number of cadavers to 30 and vary the number of PMI days from 0 to 50 days of observation. We also vary the presence of the covariate associated with the effect. For the ``Vultures'' effect, we consider the EIG obtained when preparing the 30 cadavers with vulture scavenging ``Present'' or ``Not present''. For the ``Age at Death'' and ``Body Size'' effects, we consider the EIG when placing cadavers in each of the possible covariate levels. The results are shown in Figure \ref{fig:oed_example}, top row. Unsurprisingly, EIGs are highest when the cadavers are prepared in the state corresponding to the effect being studied (e.g., vultures present for the ``Vultures'' effect). The EIGs are also higher at longer PMI durations with rapidly diminishing returns.

Figure \ref{fig:oed_example}, bottom row, shows the posterior distribution for the effect of interest before and after the experiment with the highest EIG, assuming that the mean coefficient estimates are unchanged. Adding the 30 cases reduces the expected uncertainty about the effect for the first two effects. For the ``Body size'' effect, however, the posterior distribution is nearly unchanged by the experiment. Relative to the information already present in the geoFOR dataset, the observation of 30 more emaciated cadavers is not likely to give more information about the effect of emaciation on desiccation. This observation is consistent with the relatively high precision of the posterior coefficient before the experiment.

A typical experimental design heuristic is to represent as much variation as possible in the dimensions of interest to increase the signal-to-noise ratio. This common approach assumes that the effects of interest will be measured from scratch using the experimental dataset. Using the EIG formalism, however, the effects are measured from the experimental data \textit{and} the geoFOR dataset. Heuristically, it is more reasonable to design experiments that complement the information already available in geoFOR. EIG makes this concept more precise by considering the likely outcome of the hypothetical experiment and the current state of knowledge about the effects of interest.

\section{Conclusion}

This study presented a Bayesian probabilistic model for human decomposition based on the extensive forensic taphonomy data available in the geoFOR dataset. Our model explicitly represents a wide range of decomposition mechanisms, allowing for a nuanced understanding of how various environmental and individual factors influence the progression of decomposition. By leveraging the 2,529 cases in the geoFOR dataset, we could estimate model effects and present a comprehensive set of findings related to the decomposition mechanisms. We quantified our model's performance in predicting decomposition characteristics and estimating PMI, demonstrating strong capabilities in both tasks. Additionally, we illustrated how the model can inform the design of future decomposition experiments through EIG analysis. Our approach represents a significant step forward in the field of forensic taphonomy and human decomposition research. Moving beyond simplistic measures like total body score and accumulated degree days, we have created a more detailed and flexible framework for understanding and predicting human decomposition.

While our model shows promise, viewing it as a foundation for future research rather than a final product is essential. We encourage the forensic science community to build upon this work, refining the model structure, incorporating additional variables, and validating its performance across diverse environments and populations. The optimal experimental design capabilities demonstrated in this paper offer a pathway for efficiently expanding our knowledge of decomposition processes. As we continue to refine our understanding of human decomposition through targeted research and improved modeling techniques, we move closer to more accurate and reliable PMI estimation methods, ultimately enhancing the capabilities of forensic investigators and the justice system.

\bibliographystyle{elsarticle-num}
\bibliography{bibliography}

\end{document}


\begin{frontmatter}

    \title{Supplemental Material for ``Modeling human decomposition: a Bayesian approach"}

    \author[inst1]{D.~Hudson Smith}
\author[inst2]{Noah Nisbet}
\author[inst3]{Carl Ehrett}
\author[inst4]{Cristina I.~Tica}
\author[inst5]{Madeline M.~Atwell}
\author[inst5]{Katherine E.~Weisensee}

\affiliation[inst1]{organization={Department of Mathematical and Statistical Sciences, Clemson University},
    addressline={220 Parkway Dr.},
    city={Clemson},
    postcode={29630},
    state={SC},
    country={USA},
}
\affiliation[inst2]{organization={School of Computing, Clemson University},
    city={Clemson},
    postcode={29630},
    state={SC},
    country={USA},
}
\affiliation[inst3]{organization={Research Computing and Data, Clemson University},
    city={Clemson},
    postcode={29630},
    state={SC},
    country={USA},
}
\affiliation[inst4]{organization={Department of Anthropology and Applied Archaeology, Eastern New Mexico University},
    city={Portales},
    postcode={88130},
    state={NM},
    country={USA},
}
\affiliation[inst5]{organization={Department of Sociology, Anthropology	and Criminal Justice, Clemson University},
    city={Clemson},
    postcode={29630},
    state={SC},
    country={USA},
}

    \begin{abstract}
        This supplemental material provides additional details and derivations to support the paper, ``Modeling human decomposition: a Bayesian approach.”
    \end{abstract}

\end{frontmatter}

\section{PMI Calculation Methods}\label{sec:pmi-calc}
geoFOR collects either the date of death if known or date last known alive from the respondent. Dates are provided as ``exact" (the respondent has high confidence in the specific date), ``approximate" (the respondent is unsure of the precise date), or ``range" (the respondent provides a range representing possible dates). With the exception of ``range", PMIs are estimated based on the time difference between the date of discovery and the date of death or date last known alive. For ``range", we take the midpoint between the endpoints of the range. Table \ref{tab:calc_method} below presents the counts of each calculation method used during preprocessing.

\def\arraystretch{1.1}
\begin{table}[H]
    \centering
    \caption{Number of cases for each PMI calculation method. "Longitudinal" refers to cases at decomposition research facilities. For these cases, PMI is known exactly. The single case marked "Date of death, unknown" is a retrospective case where the precision of the recorded date is unknown.}
    \begin{tabular}{ l | l }
        \textbf{PMI calculation method} & \textbf{Count} \\\hline
        Longitudinal                    & 752            \\
        Date of death, approximate      & 534            \\
        Last known alive, exact         & 463            \\
        Last known alive, approximate   & 308            \\
        Date of death, exact            & 180            \\
        Date of death, range            & 55             \\
        Date of death, unknown          & 1
    \end{tabular}
    \label{tab:calc_method}
\end{table}
\section{Mathematical notation}

Table \ref{tab:notation} describes all of the mathematical notation when describing our decomposition models.

\begin{table}[t!]\caption{Mathematical Notation}
    \begin{center}

        \begin{tabular}{l p{12cm}}
            \toprule
            $\mathcal{D}$           & The dataset; the set of all decomposition cases.
            Each case includes the observed PMI value, covariates,
            and decomposition characteristics for the case.                                             \\
            $N$                     & Number of decomposition cases in $\mathcal{D}$.                   \\
            $n$                     & An index used to specify the decomposition case in $\mathcal{D}$.
            $n$ is a normal number ranging from $1$ to $N$.                                             \\
            $D$                     & Number of decomposition characteristics.                          \\
            $d$                     & An index used to specify the decomposition characteristic.
            $d$ is a normal number ranging from $1$ to $D$.                                             \\
            $C_d$                   & Number of covariates used when modeling
            decomposition characteristic $d$.                                                           \\
            $c$                     & An index used to specify the covariate.
            $c$ is a normal number ranging from $1$ to $C_d$.                                           \\
            $L_c$                   & The number of levels in covariate $c$. For binary covariates,
            the number is two.                                                                          \\
            $\ell$                  & An index used to specify the level of the covariate.
            $\ell$ is a normal number ranging from $1$ to $L_c$.                                        \\
            $\supp{t}{n}$           & The postmortem interval period in days
            for case $n$.                                                                               \\
            $\supp{x}{n}_{c}$       & Covariate $c$ for case $n$.
            The values are a mixture of categorical and binary variables.                               \\
            $\supp{\mathcal{X}}{n}$ & Set of all covariates for case $n$.                               \\
            $\supp{y}{n}_{d}$       & The value of decomposition characteristic $d$ for case $n$.
            The possible values are 0 or 1
            indicating the absence or presence
            of the characteristics, respectively.                                                       \\
            $\supp{\mathcal{Y}}{n}$ & Set of all decomposition characteristics for case $n$.            \\
            $\supp{p}{n}_{d}$       & The probability of observing decomposition
            characteristics $d$ for case $n$.
            The values are real numbers between 0 and 1.                                                \\
            $\beta_{dc\ell}$        & The global regression coefficients relating
            categorical level $\ell$ of covariate $\supp{x}{n}_{c}$
            to $\supp{p}{n}_{d}$.
            For binary covariates, $\ell$ takes values 1 or 2
            and $\beta_{dc1}\equiv 0$.
            For categorical covariates, $\beta_{dc\ell^*}\equiv 0$,
            where $\ell^*$ is the reference level.                                                      \\
            $\beta_{d0}$            & The global regression coefficients relating $\supp{T}{n}$
            to $\supp{p}{n}_{d}$.                                                                       \\
            $\gamma_{d}$            & The log-odds of observing $d$ at $\supp{T}{n}=0$.                 \\
            $\mathcal{C}$           & The set of all unknown coefficients in the model.                 \\
            $\supp{B}{n}_d$         & The total rate coeffocient
            for decomposition characteristic $d$ and case $n$.                                          \\
            \bottomrule
        \end{tabular}
    \end{center}
    \label{tab:notation}
\end{table}

\section{Prior specifications}\label{sec:priors}
Our inference procedure requires that we specify prior distributions for all of the unknown coefficients in $\mathcal{C}$. We use the following prior specifications:
\begin{eqnarray}
    \beta_{dc\ell} &\sim& \mathrm{Normal}(0,2) \nonumber \\
    \beta_{d0} &\sim& \mathrm{Normal}(0,2) \nonumber \\
    \gamma_d &\sim& \mathrm{Normal}(-2,2)
    \label{eq:priors}
\end{eqnarray}
for all $d$, $c$, and $\ell$. These priors are moderately uninformative. We choose a negative mean for $\gamma_d$ because the log-odds of observing most decomposition characteristics on the date of death should be small. Future work could explore the benefits of using stronger prior specifications based on detailed knowledge about the mechanisms of decomposition.

\section{PMI inference details}\label{sec:pmi}
This sections describes our method for computing $p(t|\mathcal{X},\mathcal{Y},\mathcal{C})$. This distribution appears when performing posterior PMI prediction as in Eq.~\eqref{main-eq:pmi-infer} from the main paper. Applying Bayes rule and using the appropriate independence relations for the posterior predictive distribution, we find
\begin{equation}\label{eq:tposterior}
    p(t|\mathcal{X}, \mathcal{Y}, \mathcal{C}) = \frac{p(\mathcal{Y} | t, \mathcal{X}, \mathcal{C}) p(t)} {p(\mathcal{Y} | \mathcal{X}, \mathcal{C})}
\end{equation}
The first term in the numerator can be computed directly using the likelihood in Eq.~\eqref{main-eq:liklihood} from the main paper. Rather than place a prior on $t$ directly, we perform a change of variables $\tau = \log(1 + t)$ and place the following prior on $\tau$:
\begin{equation}
    \tau \sim \mathrm{Normal}(2.33, 1.53).
\end{equation}
The numeric values of the mean and standard deviation are estimated from the geoFOR data.
The term in the denominator of Eq.~\eqref{eq:tposterior} can be explicitly computed from our likelihood function:
\begin{equation}
    p(\mathcal{Y} | \mathcal{X}, \mathcal{C}) = \int_0^\infty d\tau p(\mathcal{Y} |\tau, \mathcal{X}, \mathcal{C}) p(\tau).
\end{equation}
We estimate this integral numerically when performing posterior PMI inference.

\section{Mathematical details for Expected Information Gain}\label{sec:eig}

This section uses independent notation from the main paper. All of the symbols are defined in the context of Section \ref{sec:eig} and should not be confused with the symbol definitions in Table \ref{tab:notation}.

\subsection{Derivation of exact EIG result}
In this section we detail our procedure for estimating the expected information gain. We first derive an exact expression for EIG in terms of probabilities that can be computed given a generative model, such as our model for decomposition. Specifically, we express EIG in terms of the conditional distribution $p(y|\Theta, \Phi, d)$, where $\Theta$ are the OED target variables and $\Phi$ are all remaining variables, which we refer to as {\it nuisance variables}. In this derivation, we assume that the latent random variables $\Theta$ and $\Phi$ are independent of the experimental design as is true for all models we define in this paper.

\begin{subequations}
    \begin{align}
        \mathrm{EIG}(d)
         & = E_{p(y|d)}\left[H(\Theta) - H(\Theta|y, d)\right] \label{eq:eig_1} \\
         & = -E_{p(y|d)}\left[
            E_{p(\Theta)}\log(p(\Theta)) -
            E_{p(\Theta|y,d)}\log(p(\Theta|y,d))
        \right] \label{eq:eig_2}                                                \\
         & = -E_{p(\Theta)}\log(p(\Theta))
        + E_{p(y, \Theta|d)}\log(p(\Theta|y,d)) \label{eq:eig_3}                \\
         & = -E_{p(\Theta)}\log(p(\Theta))
        + E_{p(y, \Theta|d)}\left[
            \log(p(y|\Theta,d)) +
            \log(p(\Theta|d)) -
            \log(y|d)
        \right] \label{eq:eig_4}                                                \\
         & = E_{p(y, \Theta|d)}\left[
            \log(p(y|\Theta,d)) -
            \log(p(y|d))
        \right] \label{eq:eig_5}                                                \\
         & = E_{p(y, \Theta|d)}\left[
            \log\left(\int_{-\infty}^{\infty} p(y, \Phi |\Theta,d) d\Phi \right) -
            \log\left(\int_{-\infty}^{\infty} p(y,\Phi,\Theta'|d)d\Phi d\Theta' \right)
        \right] \label{eq:eig_6}                                                \\
         & = E_{p(y, \Theta|d)}\left[
            \log\left(E_{p(\Phi|\Theta)}[p(y|\Theta,\Phi, d)]\right) -
            \log\left(E_{p(\Theta', \Phi)}[p(y|\Theta', \Phi, d)]\right)
            \right] \label{eq:eig_7}
    \end{align} \label{eq:exact_eig}
\end{subequations}

We justify these equalities in detail as follows:
\begin{itemize}
    \item Equation \eqref{eq:eig_1} is the definition of EIG.
    \item Equation \eqref{eq:eig_2} inserts the definition of entropy.
    \item Equation \eqref{eq:eig_3}, first term, performs the expectation over $y$. The second term uses
          \begin{equation}
              E_{p(y|d)}E_{p(\Theta|y,d)}[\cdot] = E_{p(y,\Theta|d)}[\cdot].
          \end{equation}
    \item Equation \eqref{eq:eig_4} applies Bayes theorem to $p(\theta|y,d)$ and expands the logarithm.
    \item Equation \eqref{eq:eig_5} assumes $p(\Theta|d) = p(\Theta)$. This conditional independence holds for the set of models and designs that we consider.
    \item Equation \eqref{eq:eig_6} inserts the definition of the marginal distribution.
    \item Finally, Equation \eqref{eq:eig_7} applies the chain rule for joint probabilities along with the definition of expectation.
\end{itemize}
In the last two expressions, we use $\Theta'$ for the inner expectation involving $\Theta$ to avoid confusion. However, $\Theta$ and $\Theta'$ correspond to the same random variables in our model. Note that, in the final result, the first term depends on an expectation over the conditional distribution $p(\Phi|\Theta)$ while second term depends on the joint distribution $p(\Theta', \Phi)$.

\subsection{EIG estimation}
For most models, including those used in our study, Eq.~\eqref{eq:exact_eig} cannot be computed exactly. Here we derive our Monte-Carlo approximation scheme. We start by approximating the expectation values using samples from our fit model:
\begin{equation}
    \begin{split}
        \mathrm{EIG}(d) \approx \frac{1}{N}\sum_{n=1}^N \left[
            \log \left(\frac{1}{M}\sum_{m=1}^Mp(y^{(n)}|\Theta^{(n)},\Phi^{(m, n)}, d)\right)\right. \\
            \left.-\log \left(\frac{1}{M'}\sum_{m'=1}^{M'}p(y^{(n)}|\Theta^{(m')}, \Phi^{(m')}, d)\right)\right]
    \end{split}
    \label{eq:eig_approx}
\end{equation}
where $N$, $M$, and $M'$ are the numbers of samples drawn according to
\begin{subequations}
    \begin{eqnarray}
        y^{(n)},\Theta^{(n)} &\sim& p(y,\Theta|d) \label{eq:samples_1}\\
        \Theta^{(m')}, \Phi^{(m')} &\sim& p(\Theta, \Phi) \label{eq:samples_2} \\
        \Phi^{(m,n)} &\sim& p(\Phi|\Theta^{(n)}) \label{eq:samples_3}
    \end{eqnarray}
    \label{eq:samples}
\end{subequations}
Equation \eqref{eq:eig_approx} becomes exact in the limit of infinite samples.

Our decomposition models allow for direct computation of the log-liklihoods $l(y|\Theta, \Phi, d) = \log(p(y|\Theta, \Phi, d))$. For numerical stability, we re-express our estimate of EIG in terms of these liklihoods:
\begin{eqnarray}
    \mathrm{EIG}(d) &\approx& \frac{1}{N}\sum_{n=1}^N \left[
        \log \left(\sum_{m=1}^M \exp\left( l(y^{(n)}|\Theta^{(n)},\Phi^{(m)}, d)\right)\right)\right. \nonumber \\
        &-&\log \left(\sum_{m'=1}^{M'}\exp\left(l(y^{(n)}|\Theta^{(m')}, \Phi^{(m')}, d)\right)\right)
        \left.+\log \left(\frac{M'}{M}\right)\right]
    \label{eq:eig_approx_log}
\end{eqnarray}
The last term inside the sum can be ignored since it is independent of the design $d$. The other terms both involve taking the logarithm of a sum of exponential factors. We use the "log-sum-exp trick" to perform these operations in a numerically stable fashion.

\subsection{Sampling for EIG estimator}
We generate the samples in Eq.~\eqref{eq:samples} as follows: for \eqref{eq:samples_1}, we directly sample from our fitted decomposition model for the joint distribution of $y$, $\Theta$, and $\Phi$ and simply discard the unneeded nuisance variables. In practice, it is convenient to take $N$ samples of $\Theta$ and $\Phi$ from our MCMC fitting procedure, plug these into the model to generate $y$, and then discard $\Phi$. Similarly, for \eqref{eq:samples_2} we simply take $M'$ samples of $\Theta$ and $\Phi$ from our MCMC fit.

Equation \eqref{eq:samples_3} requires more careful treatment. In our prior model, $\Phi$ and $\Theta$ are independent. However, the fit procedure introduces dependencies, making sampling $\Phi|\Theta$ more complicated. Based on the fact that we use normal priors for $\Theta$ and $\Phi$ and the observation that the posterior distributions appear to be approximately normal, we approximate the posterior distribution $p(\Theta, \Phi)$ using a multivariate normal distribution with parameters:
\begin{eqnarray}
    \mu &=&
    \begin{bmatrix}
        \mu_\Phi \\
        \mu_\Theta
    \end{bmatrix} \nonumber \\
    \Sigma &=&
    \begin{bmatrix}
        \Sigma_\Phi           & \Sigma_{\Phi\Theta} \\
        \Sigma_{\Phi\Theta}^T & \Sigma_\Theta
    \end{bmatrix}.
    \label{eq:normal}
\end{eqnarray}
The conditional distribution, $p(\Phi|\Theta) = \frac{p(\Phi, \Theta)}{p(\Theta)}$,
is then also a multivariate normal distribution with parameters,
\begin{eqnarray}
    \mu_{\Phi|\Theta} &=& \mu_\Phi + \Sigma_{\Phi\Theta} \Sigma_\Theta^{-1}(\Theta - \mu_\Theta)\nonumber \\
    \Sigma_{\Phi|\Theta} &=& \Sigma_\Phi - \Sigma_{\Phi\Theta} \Sigma_{\Theta}^{-1} \Sigma_{\Phi\Theta}^T.
\end{eqnarray}
This allows for direct sampling of $\Phi|\Theta$ using common software libraries.

We use the expressions for the sample mean and covariance to estimate the multivariate normal parameters from our MCMC fit samples. Letting $\mathcal{T}^{(l)} = \langle\Theta^{(l)}, \Phi^{(l)}\rangle$ be the $l$-th sample of the concatenated target and nuisance variables, the estimates are
\begin{eqnarray}
    \hat\mu_i &=& \frac{1}{L} \sum_{l=1}^L \mathcal{T}^{(l)}_i \nonumber \\
    \hat\Sigma_{ij} &=& \frac{1}{L-1}\sum_{l=1}^L (\mathcal{T}^{(l)} - \hat\mu)_i(\mathcal{T}^{(l)} - \hat\mu)_j
\end{eqnarray}
where $L$ is the number of samples. In the limit of very large $L$, the error in these estimates can be neglected in which case the dominant error in computing Eq.~\eqref{eq:eig_approx_log} comes from the normality assumption in Eq.~\eqref{eq:normal}.

\subsection{Lower-variance EIG estimator}
In the EIG estimator defined in Eq.~\eqref{eq:eig_approx_log}, sampling the outcomes $y^{(n)}$ adds to the overall variance of the estimator. If $y$ has a discrete set of possible values $y\in \mathcal Y$, we can derive a lower-variance estimator by explicitly computing the expectation over $y$ in Eq.~\eqref{eq:exact_eig}. To derive the low-variance estimator, we rewrite \eqref{eq:eig_7} in terms of an explicit sum over $y$:
\begin{equation}
    \mathrm{EIG}(d) = E_{p(\Theta, \Phi|d)}\left[\sum_{y\in \mathcal Y} p(y |\Theta, \Phi | d) f(y, \Theta, d)\right],
\end{equation}
where $f(y, \Theta, d)$ is the quantity in square brackets in \eqref{eq:eig_7}. We correspondingly modify our Monte-Carlo estimator, Eq.~\eqref{eq:eig_approx_log}. Using samples from the fit model, we approximate
\begin{align}
    \mathrm{EIG}(d) \approx \frac{1}{N}\sum_{n=1}^N & \left[\sum_{y\in \mathcal Y} p(y|\Theta^{(n)}, \Phi^{(n)}, d) \left\{ \log\left(\sum_{m=1}^{M}p(y|\Theta^{(n)}, \Phi^{(m,n)}, d)\right)\right. \right. \nonumber \\
                                                    & - \left.\left. \log\left(\sum_{m'=1}^{M'}p(y|\Theta^{(m')}, \Phi^{(m')}, d)\right) + \log\left(\frac{M'}{M}\right)\right\} \right]
\end{align}
with the sampling rules defined in Eq.~\eqref{eq:samples}. Note that $\Theta^{(n)}, \Phi^{(n)}$ are sampled according to Eq.~\eqref{eq:samples_2}. As before, we can apply the log-sum-exp trick arriving at the final form of the low-variance estimator:
\begin{align}
    \mathrm{EIG}(d) \approx \frac{1}{N}\sum_{n=1}^N & \left[\sum_{y\in \mathcal Y} p(y|\Theta^{(n)}, \Phi^{(n)}, d) \left\{ \log\left(\sum_{m=1}^{M}\exp \left(l(y|\Theta^{(n)}, \Phi^{(m,n)}, d)\right)\right)\right. \right. \nonumber \\
                                                    & - \left.\left. \log\left(\sum_{m'=1}^{M'}\exp \left(l(y|\Theta^{(m')}, \Phi^{(m')}, d)\right)\right) + \log\left(\frac{M'}{M}\right)\right\} \right]
\end{align}

\section{EIG toy model example}
We provide a simple Bayesian regression example to illustrate the concept of EIG and to validate our estimation strategy.

\subsection{Toy model specification}
Consider a simple linear regression model with a single covariate $x$ and a single response variable $y$ and with Normal priors on the slope, $\theta$, and the intercept, $\phi$. This model can be written as
\begin{equation}
    p(y,\theta,\phi|x) = p(\theta)p(\phi)p(y|x, \theta, \phi)\label{eq:toy_factorization}
\end{equation}
with
\begin{eqnarray}
    \theta &\sim& \mathrm{Normal}(0,\sigma_\theta) \nonumber \\
    \phi &\sim& \mathrm{Normal}(0,\sigma_\phi) \nonumber \\
    y|\theta,\phi &\sim& \mathrm{Normal}(\theta x + \phi,\sigma).
    \label{eq:toy_dists}
\end{eqnarray}
We assume that the parameters $\sigma_\theta$, $\sigma_\phi$, and $\sigma$ are known.

\subsection{Exact EIG computation}
For this simple model, the EIG can be computed exactly. For this computation, we identify the design, $d$, with the value of the $x$ covariate. Using Equations \eqref{eq:toy_factorization} and \eqref{eq:toy_dists} in Equation \eqref{eq:eig_5}, the exact EIG expressions are
\begin{subequations}
    \begin{eqnarray}
        \mathrm{EIG}_\theta(x) &=& \frac{1}{2}\log\left(1 + \frac{(\sigma_\theta x)^2}{\sigma^2 + \sigma_\phi^2} \right) \label{eq:toy_eig_theta_a}\\
        \mathrm{EIG}_\phi(x) &=& \frac{1}{2}\log\left(1 + \frac{\sigma_\phi^2}{\sigma^2 + (\sigma_\theta x)^2} \right),
        \label{eq:toy_eig_theta_b}
    \end{eqnarray}
    \label{eq:toy_exact}
\end{subequations}
where $\mathrm{EIG}_\theta(x)$ is the expected information gain for the slope, $\theta$, and, similarly, $\mathrm{EIG}_\phi(x)$ is expected information gain for the intercept, $\phi$. Figure \ref{fig:toy_eig} plots these expressions as a function of $x$ for $\sigma=0.5$ and $\sigma_\theta = \sigma_\phi = 1$. For $\theta$, the EIG is zero when $x=0$ as it must be since $\theta$ does not appear in the model when $x=0$. The EIG grows monotonically for larger $x^2$ as observations of $y$ at these points tend to constrain the value of the slope more effectively. Inversely, when $x=0$ the effect of the intercept, $\phi$, is isolated and the corresponding EIG is maximal. The EIG for $\phi$ approaches zero for larger $x^2$ since variations in $y$ are more easily explained by changes to the slope for large values of $x^2$.

\subsection{EIG estimation}
\begin{figure}[H]
    \centering
    \includegraphics[width=\textwidth]{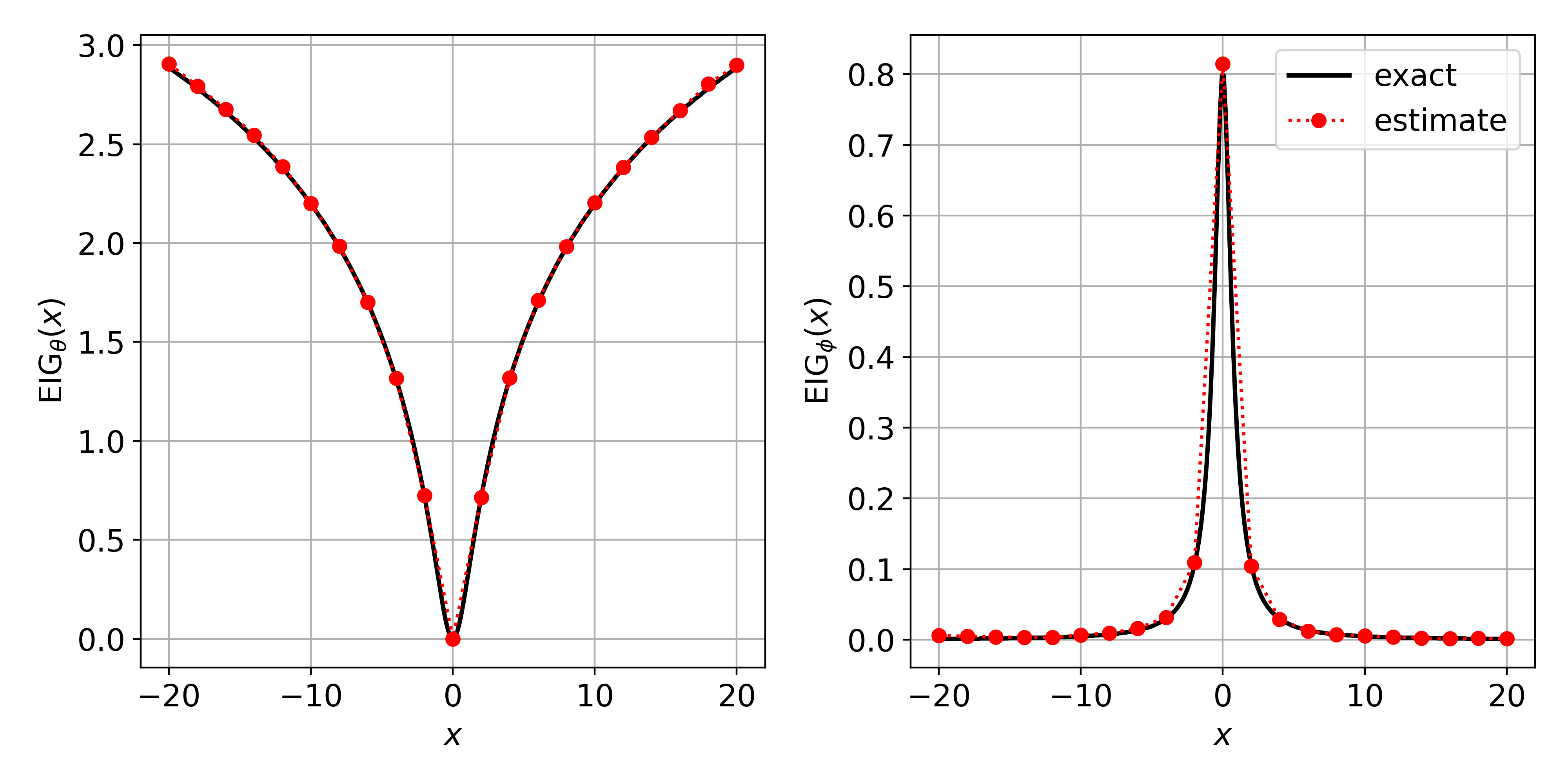}
    \caption{Comparison of the exact EIG for the toy model in Eq.~\eqref{eq:toy_exact} with the EIG computed using our estimator in Eq.~\eqref{eq:eig_approx_log}. The estimator is in good agreement with the exact result. The left panel shows the expected information gain for the slope parameter $\theta$. The right panel shows the expected information gain for the intercept parameter $\phi$.}
    \label{fig:toy_eig}
\end{figure}

We also estimate the EIG for the toy model using Eq.~\eqref{eq:eig_approx_log} and Eq.~\eqref{eq:samples}. It is not necessary to separately sample $\theta^{(m')}$ using Eq.~\eqref{eq:samples_3} since, for the toy model, $\theta$ and $\phi$ are independent. We use $N=10^4$ and $M=M'=5\times10^3$ for the estimation. The results are shown in Figure \ref{fig:toy_eig}. We find that the estimator is in good agreement with the exact result.

\section{ROC Curves}\label{sec:roc_curves}
\begin{figure}[H]
    \centering
    \includegraphics[scale=0.6]{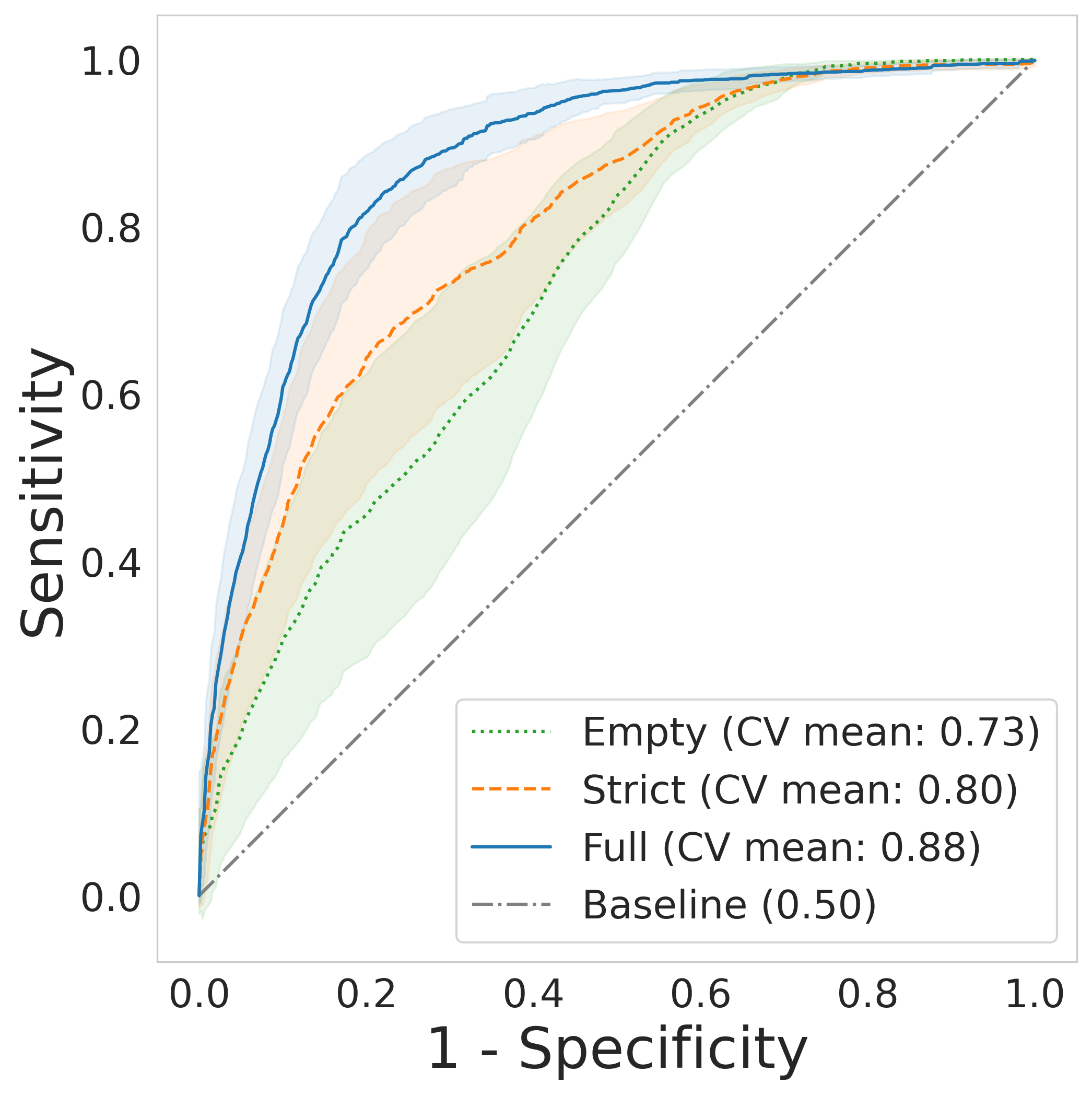}
    \caption{Receiver operating characteristic (ROC) curves for each model variant. Each model's ROC curve is an average over the 24 decomposition characteristics and the five cross validation folds. We interpolate points along each fold to construct 95\% confidence intervals. The dotted line indicates a model that randomly predicts decomposition characteristics. A perfect model has an area under the curve of one.}
    \label{fig:roc_curves}
\end{figure}